\documentclass[journal]{IEEEtran}
\IEEEoverridecommandlockouts

\usepackage{cite}
\usepackage{comment} 
\usepackage{amsmath,amssymb,amsfonts}
\usepackage{graphicx}
\usepackage{textcomp}
\usepackage{xcolor}
\usepackage{float}
\usepackage{graphicx}
\usepackage{setspace}
\usepackage{amssymb}
\usepackage[english]{babel}
\usepackage{acronym}
\usepackage{units}
\usepackage{multirow}
\usepackage{url}
\usepackage{sidecap}
\usepackage{todonotes}
\usepackage{multirow}
\usepackage{algpseudocode}
\usepackage{algorithm}
\usepackage{subcaption}

\algdef{SE}[EVENT]{Event}{EndEvent}[1]{\textbf{upon event}\ #1\ \algorithmicdo}{\algorithmicend\ \textbf{event}}%
\algtext*{EndEvent}

\newcommand{\refeq}[1]{{Eq.~(\ref{#1})}}

\newcommand{\reffig}[1]{{Fig.~\ref{#1}}}
\newcommand{\reftab}[1]{{Tab.~\ref{#1}}}

\newcommand{\refsec}[1]{{Sec.~\ref{#1}}}

\newcommand{\pd}[2]{\frac{\partial #1}{\partial #2}}
\newcommand{\ptd}[2]{\frac{\mathrm{d} #1}{\mathrm{d}#2}}

\normalsize

\acrodef{AER}[AER]{Address Event Representation}
\acrodef{ANN}[ANN]{Artificial Neural Network}
\acrodef{BL}[BL]{Bit Line}
\acrodef{BP}[BP]{Back-Propagation}
\acrodef{BPTT}[BPTT]{Back-Propagation-Through-Time}
\acrodef{DPI}[DPI]{Differential-Pair Integrator}
\acrodef{EPSC}[EPSC]{Excitatory Post-Synaptic Current}
\acrodef{EPSP}[EPSP]{Excitatory Post--Synaptic Potential}
\acrodef{GPU}[GPU]{Graphical Processing Unit}
\acrodef{LIF}[LI\&F]{Leaky Integrate \& Fire}
\acrodef{NC}[NC]{Neuromorphic Core}
\acrodef{PC}[PC]{Processing Core}
\acrodef{PCM}[PCM]{Phase Change Memory}
\acrodef{RCA}[RCA]{Resistive Crossbar Array}
\acrodef{RTRL}[RTRL]{Real-Time Recurrent Learning}
\acrodef{SRM}[SRM]{Spike Response Model}
\acrodef{SL}[SL]{Source Line}
\acrodef{SNN}[SNN]{Spiking Neural Network}
\acrodef{STDP}[STDP]{Spike Time Dependent Plasticity}
\acrodef{TIA}[TIA]{Transimpedence Amplifier}
\acrodef{SG}[SG]{Surrogate Gradient}
\acrodef{VMM}[VMM]{Vector Matrix Multiplication}
\acrodef{VWBump}[VWBump]{Variable Width Bump}
\acrodef{WTA}[WTA]{Winner--Take--All}
\acrodef{WL}[WL]{Word Line}

\begin{document}

\title{On-Chip Error-triggered Learning of Multi-layer Memristive Spiking Neural Networks}

\author{ \IEEEauthorblockN{ Melika Payvand$^{*}$, Mohammed E. Fouda$^{*}$, Fadi Kurdahi, Ahmed M. Eltawil, and Emre O. Neftci\\
$^*$These authors contributed equally to this work.}
\thanks{Melika Payvand is with Institute of neuroinformatics, University of Zurich and ETH Zurich, Zurich, Switzerland. }
\thanks{Mohammed Fouda and Fadi Kurdahi are with Electrical Engineering and Computer Science Dept., UC Irvine, Irvine, CA 92697-2625 USA.} 
\thanks{Ahmed Eltawil is with King Abdullah University of Science and Technology (KAUST), Thuwal, Saudia Arabia and with  Electrical Engineering and Computer Science Dept., UC Irvine, Irvine, CA 92697-2625 USA.}
\thanks{Emre Neftci is with Dept of Cognitive Sciences and with Dept. of Computer Science, UC Irvine, Irvine, CA 92697-2625 USA}
\thanks{This work was supported by the National Science Foundation under grant 1652159 and 1823366 (EON).}
}

\maketitle

\begin{abstract}
Recent breakthroughs in neuromorphic computing show that local forms of gradient descent learning are compatible with \acp{SNN} and synaptic plasticity.
Although \acp{SNN} can be scalably implemented using neuromorphic VLSI, an architecture that can learn using gradient-descent \emph{in situ} is still missing. In this paper, we propose a local, gradient-based, error-triggered learning algorithm with online ternary weight updates. The proposed algorithm enables online training of multi-layer \acp{SNN} with memristive neuromorphic hardware showing a small loss in the performance compared with the state-of-the-art. We also propose a hardware architecture based on memristive crossbar arrays to perform the required vector-matrix multiplications. The necessary peripheral circuitry including presynaptic, post-synaptic and write circuits required for online training, have been designed in the subthreshold regime for power saving with a standard 180\,nm CMOS process. 
\end{abstract}

\vspace{-0.2cm}
\section{Introduction}
The implementation of learning dynamics as synaptic plasticity in neuromorphic hardware can lead to highly efficient, lifelong learning systems \cite{Davies_etal18_loihneur,frenkel20180,Qiao_etal15_recoon-l,Friedmann_etal17_demohybr}.
While gradient Backpropagation (BP) is the workhorse for training nearly all deep neural network architectures, gradients' computation involves information that is not spatially and temporally local \cite{Baldi_etal17_learmach}. 
This non-locality can result in a large area overhead for routing which makes it very expensive to implement in neuromorphic hardware \cite{Zenke_Neftci20_brailear}.
Recent work addresses this problem using \ac{SG}, local learning and an approximate forward-mode differentiation \cite{Neftci_etal19_surrgrad,Kaiser_etal20_synaplas,Zenke_Neftci20_brailear}.
SGs define a differentiable surrogate network used to compute weight updates in the presence of non-differentiable spiking non-linearities \cite{Zenke_Ganguli17_supesupe,Neftci_etal19_surrgrad}.
Local loss functions enable updates to be made in a spatially local fashion \cite{Kaiser_etal20_synaplas}.
The approximate forward mode differentiation is a simplified form of \ac{RTRL} \cite{Williams_Zipser89_learalgo} that enables online learning using temporally local information \cite{Neftci_etal19_surrgrad}.
The result is a learning rule that is both \emph{spatially and temporally} local, which takes the form of a three-factor synaptic plasticity rule. 

The \ac{SG} approach reveals, from first principles, the mathematical nature of the three factors, enabling thereby a distributed and online learning dynamic.

In this article, we design a hardware architecture, learning circuit and learning dynamics that meet the realities of circuit design and mathematical rigor.
Our resulting learning dynamic is an error-triggered variation of gradient-based three factor rules that is suitable for efficient implementation in \acp{RCA}.
Conventional backpropagation schemes requires separate training and inference phases which is at odds with learning efficiently on a physical substrate \cite{Baldi_etal17_learmach}.
In the proposed learning dynamic, there is no backpropagation through the main branch of the neural network. Consequently, the learning phase can be naturally interleaved with the inference dynamics and only elicited when a local error is detected. 
Furthermore, error-triggered learning leads to a smaller number of parameter updates necessary to reach the final performance, which positively impacts the endurance and energy efficiency of the training, by factor up to $88\times$.

\acp{RCA} present an efficient implementation solution for Deep Neural Networks (DNNs) acceleration. 
The \ac{VMM}, which is the corner-stone of DNNs, is performed in one step compared to $O(N^2)$ steps for digital realizations where $N$ is the vector dimension. 
A surge of efforts focused on using \acp{RCA} for \acp{ANN} such as \cite{li2018efficient,bayat2018implementation,wang2019reinforcement,xia2019memristive} but comparatively few work utilize \acp{RCA} for spiking neural networks trained with gradient-based methods \cite{Neftci_etal19_surrgrad,Zenke_Ganguli17_supesupe,Shrestha_Orchard18_slayspik}. 
Thanks to the versatility of our proposed algorithm, \acp{RCA} can be fully utilized with suitable peripheral circuits. We show that the proposed learning dynamic is particularly well suited for the \ac{RCA}-based design and performs near or at deep learning proficiencies with a tunable accuracy-energy trade-off during learning.

\vspace{-0.2cm}
\subsection{State-of-the-art and Related Work}

Learning in neuromorphic hardware can be performed as off-chip learning, or using a hardware-in-the-loop training, where a separate processor computes weight updates based on analog or digital states \cite{Nandakumar_etal20_mixedeep,Friedmann_etal17_demohybr}.
While these approaches lead to performance that is on par with conventional deep neural networks \cite{Cramer_etal20_traispik,Wozniak_etal20_deeplear}, they do not address the problem of learning scalably and online.

In a physical implementation of learning, the information for computing weight updates must be available at the synapse. 
One approach is to convey this information to the neuron and synapses. However, this approach comes at a significant cost in wiring, silicon area, and power. 
For an efficient implementation of on-chip learning, it is necessary to design an architecture that naturally incorporates the local information at the neuron and the synapses.
For example, Hebbian learning or its spiking counterpart \ac{STDP}, depend on pre-synaptic and post-synaptic information and thus satisfy this requirement. 
Consequently, many existing on-chip learning approaches focus on their implementation in the forms of unsupervised and semi-supervised learning ~\cite{kuzum2012TED,wang2015FrontNeurosci,covi2016FrontNeurosci,pedretti2017SciRep,prezioso2018NatComm}. 
There have also been efforts in combining CMOS and memristor technologies to design supervised local error-based learning circuits using only one network layer by exploiting the properties of memristive devices \cite{Payvand_etal19_neursyst,dalgaty_etal2019_IPneuron,dalgaty_etal_2019_AIP,nair2015gradient,nair2017differential}. 
However, these works are limited in learning static patterns or shallow networks. 

In this work, we target the general multi-layer learning problem by taking into account the neural dynamics and multiple layers.
Currently, Intel Loihi research chip, Spinnaker 1 and 2, and the Brainscales-2 have the ability to implement a vast variety of learning rules \cite{Davies_etal18_loihneur,yan_etal2019,furber2014spinnaker}.
Spinnaker and Loihi are both research tools that provide a flexible programmable substrate that can implement a vast set of learning algorithms. 
This is achieved at the cost of more power and chip area consumption. 
For example, Loihi and Spinnakers's flexibility is enabled by three embedded x86 processor cores, and Arm cores, respectively. 
The plasticity processing unit used in Brainscales-2 is a general-purpose processor for computing weight updates based on neural states and extrinsic signals.
Although effective for inference stages, the learning dynamics do not break free from the conventional computing methods or use high precision processors and a separate memory block. 
In addition to requiring large amounts of memory to implement the learning, such implementations are limited by the von-Neumann bottleneck and is thus power hungry due to shuttling the data between the memory and the processing units.

In \cite{payvand2020error}, we presented the concept of error-triggered learning and a circuit implementation. This paper significantly extends the theory, {system architecture} and circuits of that work to improve scalability, area and power. 
The contributions of this paper are summarized as follows:
\begin{itemize}
    \item We extend the error-triggered learning algorithm to make the learning fully ternary to suit the targeted memristor-based \ac{RCA} hardware. 
    \item We propose a complete and novel hardware architecture that enables asynchronous error-triggered updates according to the algorithm.
    \item We propose an implementation of the neuromorphic core, including memristive crossbar peripheral circuits, update circuitry, pre-and post-synaptic circuits.   
\end{itemize}

This paper is organized as follows: \refsec{sec:et} introduces the error-triggered learning algorithm. \refsec{sec:sim_res}  discusses large scale simulation experiments {performed using PyTorch \cite{Paszke_etal17_autodiff}}\footnote{Code is available at \url{https://github.com/nmi-lab/decolle-public/}}. The hardware architecture is presented in \refsec{sec:hw_arch}. Then, the implementation of the inference and training circuits are presented in  \refsec{sec:neurocore} {and results of circuit simulations using Cadence Spectre are reported.} \refsec{sec:Discussion} discusses the limitation and potential of the proposed hardware and algorithm. Finally, the conclusion and future works are given.

\section{Error-triggered Learning Algorithm}\label{sec:et}
\subsection{Neural Network Model}
The proposed model consists of networks of plastic integrate-and-fire neurons. 

Here, the models are formalized in discrete-time to make the equivalence with classical artificial neural networks more explicit.
However, these dynamics can also be written in continuous-time without any conceptual changes. The neuron and synapse dynamics written in vector form are:

\begin{equation}\label{eq:lif_equations}
  \begin{split}
    U^l[t] &= W^l P^l[t] - \delta R^l[t], \quad S^l[t] = \Theta( U^l[t]) 
  \end{split}
\end{equation}
\vspace{-0.15in}
\[
  \begin{split}
    P^l[t+1] &= \alpha^l P[t] + Q^{l}[t], \\
    Q^l[t+1] &= \beta^l Q[t] + S^{l-1}[t], \\
    R^l[t+1] &= \gamma_i R^l[t] + S^{l}[t]. \\
  \end{split}
\]
where $U^{l}[t] \in \mathbb{R}^{N^l},\quad l \in [1,L]$ is the membrane potential of $N^l$ neurons at layer $l$ at time step $t$, $W^l$ is the synaptic weight matrix between layer $l-1$ and $l$, and $S^l$ is the binary output of this neuron. 
$\Theta$ is the step function acting as a spiking activation function, \emph{i.e.} ($\Theta(x)=1$ if $x\ge 0$, and $\Theta(x)=0$ otherwise).
The terms $\alpha^l$, $\gamma^l$, $\beta^l \in \mathbb{R}^{N^l}$ capture the decay dynamics of the membrane potential, the synapse and the refractory state $R^l$, respectively.
States $P^l$ describe the post-synaptic potential in response to input events $S^{l-1}$. 
States $Q^l$ can be interpreted as the synaptic dynamic state.
The decay terms are written in vector form, meaning that every neuron is allowed to have a different leak. 
It is important to take variations of the leak across neurons into account because fabrication mismatch in subthreshold implementations may lead to substantial variability in these parameters \cite{Neftci_etal11_systmeth}.

$R$ is a refractory state that inhibits the neuron after it has emitted a spike, and $\delta \in \mathbb{R}$ is the constant that controls its magnitude.
Note that \refeq{eq:lif_equations} is equivalent to a discrete-time version of a type of \ac{LIF} \cite{Kaiser_etal20_synaplas} and the \ac{SRM} with linear filters \cite{Gerstner_Kistler02_spikneur}.

The same dynamics can be written for recurrent spiking neural networks, whereby the same layer feeds into itself, by adding another connectivity matrix to each layer to account for the additional connections.
This \ac{SNN} and the ensuing learning dynamics can also be transformed into a standard binary neural network by setting all decay terms and $\delta$ to $0$, which is equivalent to replacing all $P$ with $S$ and dropping $R$ and $Q$.

\vspace{-0.2cm}
\subsection{Surrogate Gradient Learning}
\label{sec:surgrad}
Assuming a global cost function $\mathcal{L}[t]$ defined for the time step $t$, the gradients with respect to the weights in layer $l$ are formulated as three factors 
\begin{equation}\label{eq:loss}
\nabla_{W^l} \mathcal{L}[t] = 
\pd{\mathcal{L}}{S^l}
\pd{S^{l}}{U^{l}} 
\ptd{U^{l}}{W^{l}}
\end{equation}

where we have used $\frac{\mathrm{d}}{\mathrm{d}W^l}$ to indicate a total derivative, because the differentiated state may indirectly depend on the differentiated parameter $W$, and dropped the notation of the time $[t]$ for clarity.
The rightmost factor describes the change of the membrane potential changes with the weight $W^l$.
This term can be computed as ${P^l}[t] - \delta \ptd{R^l[t]}{W^l}$ for the neuron defined by \refeq{eq:lif_equations}.
Note that, as in all neural network calculus, this term is a sparse, rank-3 tensor.
However, for clarity and the ensuing  simplifications, we write it here as a vector.
The term with $R$ involves a dependence of the past spiking activity of the neuron, which significantly increases the complexity of the learning dynamics. 
Fortunately, this dependence can be ignored during learning without empirical loss in performance \cite{Zenke_Ganguli17_supesupe}. 
The middle factor is the change in spiking state as a function of the membrane potential, \emph{i.e.} the derivative of $\Theta$.
$\Theta$ is non-differentiable but can be replaced by a surrogate function such as a smooth sigmoidal or piecewise constant function \cite{Neftci_etal19_surrgrad}.
Our experiments make use of a piecewise linear function, such that this middle factor becomes the box function: $\pd{S_i}{U^l_i} := B(U^l_i) = 1$ if $u_-<U_i^l<u_+$ and $0$ otherwise. We define then $B^l$ as the diagonal matrix with elements $B(U^l_i)$ on the diagonal.
The leftmost factor describes how the change in the spiking state affects the loss.
It is commonly called the local error (or the ``delta'') and is typically computed using gradient BP.
We assume for the moment that these local errors are available and denote them $err^l$, and revisit this point in \refsec{sec:local_error}.
Using standard gradient descent, the weight updates become:

\begin{equation}\label{eq:spiking_neuron_rule}
  \Delta W^l = - \eta \nabla_{W^{l}} \mathcal{L} = - \eta\, (err^l B^l)^T {P^l},
\end{equation}
In scalar form, the rule simplifies as follows:
\begin{equation}\label{eq:spiking_neuron_rule}
  \Delta W_{ij}^l = - \eta\, err_i^l P_j^{l} \text{, if $u_-<U_i<u_+$},
\end{equation}
where $\eta$ is the learning rate.

\subsection{Error-triggered Learning}
By virtue of the chain rule of calculus, \refeq{eq:loss} reveals that the derivative of loss function in a neural network  (the first term of the equation, $\pd{\mathcal{L}}{S^l}$) depends solely on the output state $S$. 
The output state is a binary vector with $N^l$, and can naturally be communicated across a chip using event-based communication techniques with minimal overhead.
The computed errors ($\pd{\mathcal{L}}{S^l}$) are vectors of the same dimension, but are generally reals, \emph{i.e.} defined in $\mathbb{R}^{N^l}$. 
For \emph{in situ} learning, and as previously discussed, the error vector must be available at the neuron.
To make this communication efficient, we introduce a tunable threshold on the errors and encoded them using positive and negative events as follows:
\begin{equation}\label{eq:error-coding-neurons}
    E^l = sign(err^l)(|err^l|\div \theta^l),
\end{equation}
where $\theta^l \in \mathbb{R}$ is a constant or slowly varying error threshold unique to each layer $l$ and $\div$ is an integer division.
\reffig{fig:err_func_illustration} illustrates this function.
Note that in the formulation above, $E_i$ can exceed -1 and 1.
In this case, multiple updates are made.
Using this encoding, the parameter update rule written in scalar form becomes:

\begin{equation}\label{eq:binary_neuron_rule}
  \Delta W_{ij}^l = - \tilde{\eta} E^l_i B(U_i^l) P_j^{l},  
\end{equation}
where $\tilde{\eta}= \eta \theta$ is the new learning rate that subsumes the value of $\theta$. 
Thus, an update takes place on an error of magnitude $\theta$ and if $B(U_i^l)=1$.
The sign of the weight update is $-E^l_i$ and its magnitude $\tilde{\eta} P_j^{l}$. 
Provided that the layer-wide update magnitude can be modulated proportionally to $\tilde{\eta}$, this learning rule implies two comparisons and an addition (subtraction).

\subsection{Approximate gradient estimation}
When implementing the rule in memristor crossbar arrays, using analog values for $P$ would require coding its value as a number of pulses, which would require extra hardware. 
In order to avoid sampling the $P$ signal and simplify the implementation, we further discretize $P$ value to a binary signal by thresholding (using a simple comparator):
\[
\tilde{\ptd{U^{l}}{W^{l}}} = c \tilde{P},\text{ with }\tilde{P} = \Theta(P-\bar{p})
\]
where $c$ and $\bar{p}$ are constants, and $\tilde{P}$ is the binarized $P$.
This comparator is only activated upon weight updates and the analog value is otherwise used in the forward path.
Since $\tilde{P} \in \{0,1\}$, the constant $c$ can be subsumed in the learning rate $\tilde{\eta}$ and the parameter update becomes ternary $\Delta W^l_{ij} \in \{-\tilde{\eta},0,\tilde{\eta} \}$.
The full implementation of the encoder and CMOS sampler circuits are discussed in \cite{payvand2020error}.
\begin{figure}
    \centering
    \includegraphics[width=.3\textwidth]{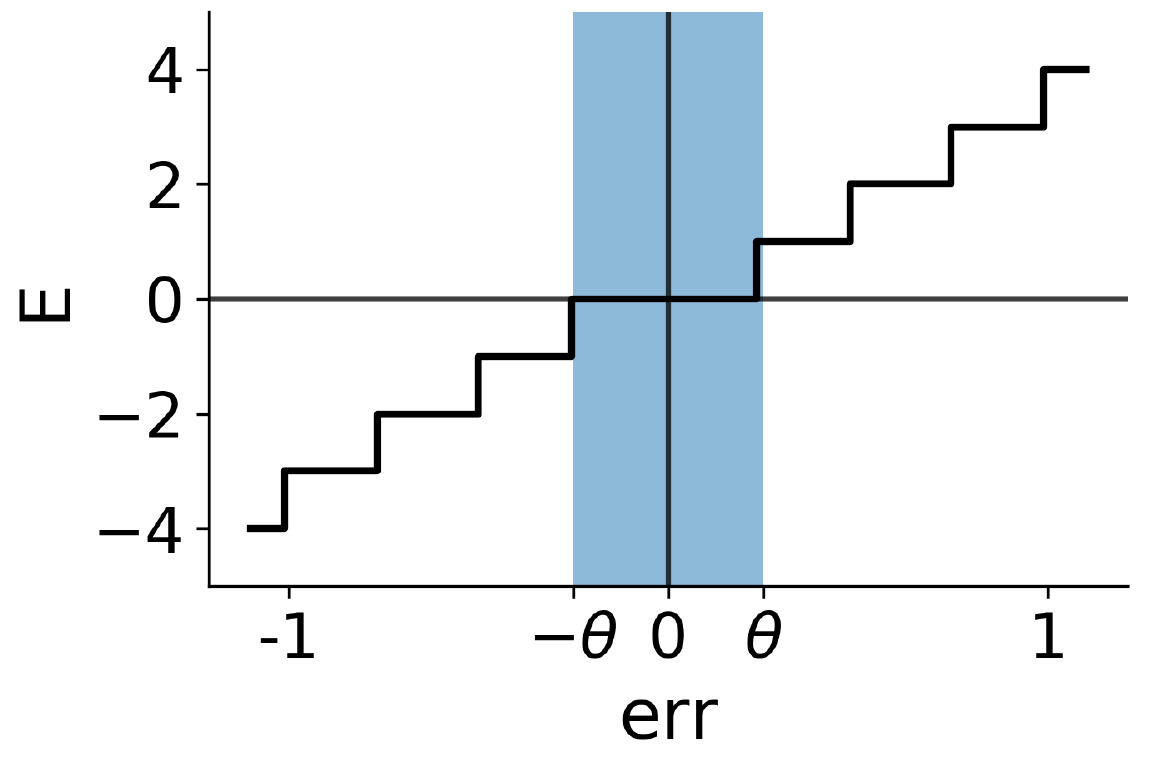}
    \caption{Illustration of the error discretization used for error-triggered learning. $E$ here is the error magnitude as a function of the real valued error $err$. Not that although the magnitude of $E$ can be larger than 1, these events are 1) rare after a few learning iterations and 2) represented as multiple ternary events.}
    \label{fig:err_func_illustration}
\vspace{-0.4cm}    
\end{figure}
 \begin{algorithm}
 \caption{Error-triggered Learning Algorithm for layer $l$}
 \label{alg:eos}
 \begin{algorithmic}
 \Require a minibatch of inputs and targets $(S^{in}, \hat{Y})$, previous weights $W$, previous $\theta$, and previous learning rate $\eta$
 \Ensure updated weights $W$, updated  parameter $\theta$.\\
 \{1.Forward propagation:\}
    \State $Q = \beta Q + S^{in}$
    \State $P = \alpha P + Q$
    \State$U\leftarrow W P - \delta R$
   \State $S \leftarrow \Theta(U)   $
 \\
 \{2. Gradient Computation:\}
  \State $\mathcal{L}= Loss(S,\hat{Y})$ 
  \State $err\leftarrow \frac{\partial{\mathcal{L} }}{\partial{U}}=B(U) \frac{\partial{\mathcal{L} }}{\partial{U}}$ 
\For {$i=1$}
\State $E_i = - sign(err_i)(|err_i|\div\theta)$
\If{$E_i^l\ne 0$} \Comment{Error triggered}
\If{$B(U_i)\ne 0$}
\For {$j=1$}
\If{$P\ge \bar{p}$}
\State $W_{ij} \leftarrow W_{ij} + \tilde{\eta} E_i $ 
\EndIf
\EndFor
\EndIf
\EndIf
\State $\theta \leftarrow$ Update$(\theta, E)$
\EndFor
\end{algorithmic}
\end{algorithm}

\begin{figure}
  \centering\includegraphics[height=.20\textheight]{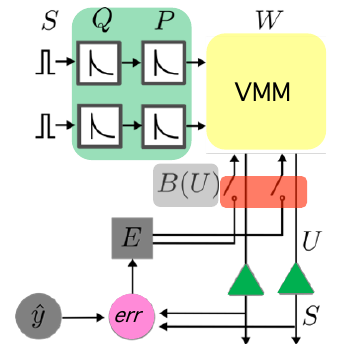}
  \centering\includegraphics[height=.20\textheight]{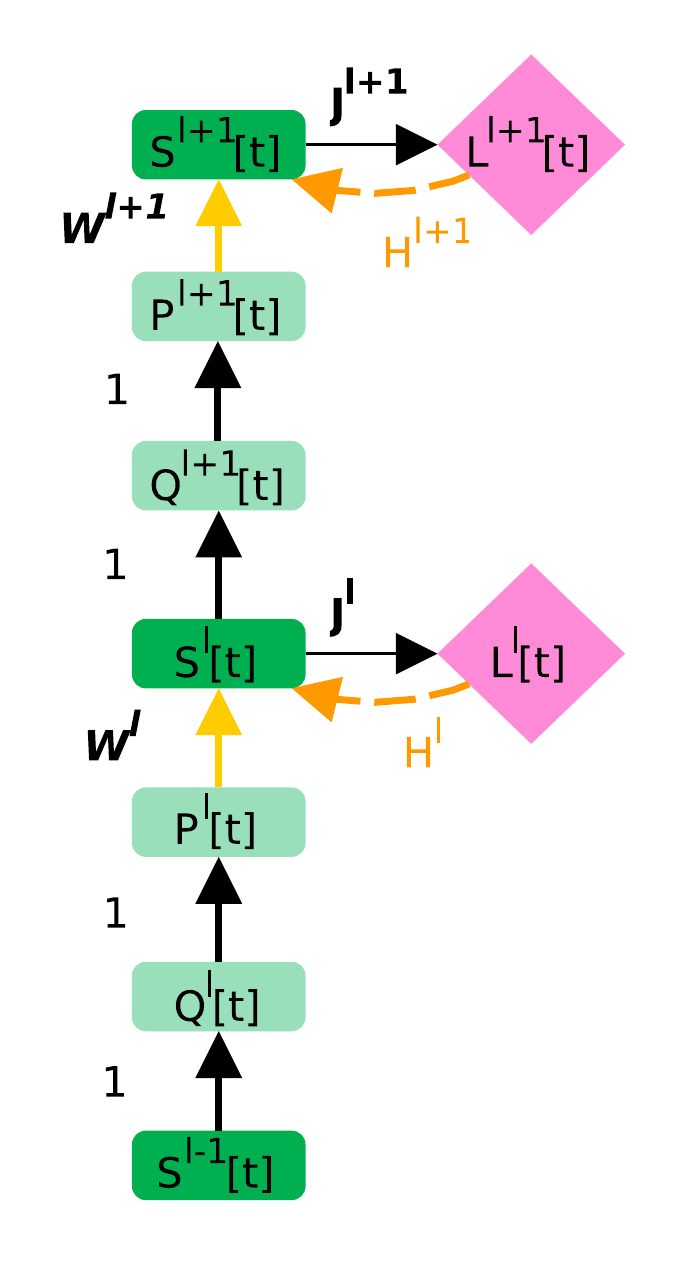}
  \caption{Architecture of the three-factor error-triggered rule (left)  and the multilayer network (right). (Left) Input spikes $S$ are integrated through $P$. The vector $P$ is then multiplied with $W$ resulting in $U$. Output spikes $S$ are then compared with local targets $\hat{Y}$ and bipolar error events $E$ are fed back to each neuron. Updates are made if $u_-<U<u_+$. Gradients are estimated using layer-local loss functions $L^l$, and are not backpropagated beyond the associated layer (Orange, dashed lines). $R$ is omitted in these diagrams to reduce clutter. }
  \label{fig:Idealarch}
\end{figure}

\subsection{Separation of Learning as Extrinsic and Intrinsic Factors}\label{sec:local_error}
Up to now, we have side-stepped the calculation of $err[n]^l$.
As discussed earlier, the factorization of the learning rule in three terms enables a natural distribution of the learning dynamics. 
The factor $E_i^l$ can be computed extrinsically, outside of the crossbar, and communicated via binary events (respectively corresponding to $E=-1$ or $E=1$) to the neurons.
A high-level architecture of the design is shown in \reffig{fig:arch}.
The computations of $E$ can be performed as part of another spiking neural network, or on a general-purpose processor (see \acs{PC} and \acs{NC} in \refsec{sec:hw_arch}).
In this article, we are agnostic to the implementation of this computation, provided that the error $E_i^l$ is projected back to neuron $i$ in one time step and that it can be calculated using $S^l$. 
Below, we describe an example of how these extrinsic errors can be computed.

\paragraph{Deep Local Losses}
If $l<L$, meaning it is not the output layer, then computing $E^l_i$ requires solving a deep credit assignment problem. 
Gradient BP can solve this, but is not compatible with a physical implementation of the neural network \cite{Baldi_etal17_learmach,Lillicrap_etal20_backbrai}, and is extremely memory intensive in the presence of temporal dynamics \cite{Williams_Zipser95_gradlear}.
Several approximations have emerged recently to solve this, such as feedback alignment \cite{Lillicrap_etal16_randsyna, Neftci_etal17_evenranda, N-kland16_direfeed}, and local losses defined for each layer \cite{Mostafa_etal18_deepsupe,Kaiser_etal20_synaplas,Nkland_Eidnes19_traineur}.
For classification, examples of local losses are layer-wise classifiers (using output labels) \cite{Mostafa_etal18_deepsupe} and supervised clustering, which can perform on par with BP in classical ML benchmark tasks \cite{Nkland_Eidnes19_traineur}.
For simplicitly, in this article we use a layer-wise local classifier using a mean-squared error loss defined as $\mathcal{L}_i^l = ||\sum_{k=1}^C (J^l_{ik} S^l_k-\hat{Y}_k)||_2$, 
where $J^l_{ik}$ is a random, fixed matrix, $\hat{Y}_k$ are one-hot encoded labels, and $C$ is the number of classes.
The gradients of $\mathcal{L}_i^l$ involve backpropagation within the time step $t$ and thus requires the symmetric transpose, $J^{l,T}$. 
If this symmetric transpose is available, then $\mathcal{L}$ can be optimized directly. 
To account for the case where $J^T$ is unavailable, for example in mixed signal systems, we train through feedback alignment using another random matrix $H^l$ \cite{Lillicrap_etal16_randsyna} whose elements are equal to $H_{ij}^l = J_{ij}^{l,T}\omega_{ij}^l$ with Gaussian distributed $\omega_{ij}^l \sim N(1,\frac12)$, where $T$ indicates transpose. 
This strategy was also used in our previous work \cite{Kaiser_etal20_synaplas}.

Using this strategy, the error can be computed with any loss function (\emph{e.g.} mean-squared error or cross entropy) provided there is no temporal dependency, \emph{i.e.} $\mathcal{L}[t]$ does not depend directly on variables in time step $t-1$ \cite{Zenke_Neftci20_brailear}. 
If such temporal dependencies exist, for example with Van Rossum spike distance, the complexity of the learning rule increases by a factor equal to the number of post-synaptic neurons \cite{Zenke_Ganguli17_supesupe}.
This increase in complexity would significantly complicate the design of the hardware. Consequently, our approach does not include temporal dependencies in the loss function.

The matrices $J^l$ and $H^l$ can be very large, especially in the case of convolutional networks.
Because these matrices are not trained and are random, there is considerable flexibility in implementing them efficiently. 
One solution to the memory footprint of these matrices is to generate them on the fly, for example using a random number generator or a hash function. 
Another solution is to define $J^l$ as a sparse, binary matrix \cite{Baldi_etal17_learmach}.
Using a binary matrix would further reduce the computations required to evaluate $err$.

The resulting learning dynamics imply no backpropagation through the main branch of the network. Instead, each layer learns individually. It is partly thanks to the local learning property that updates to the network can be made in a continual fashion, without artificial separation in learning and inference phases \cite{Kaiser_etal20_synaplas}.

\begin{figure*}
    \centering
    \includegraphics[width=.3\textwidth]{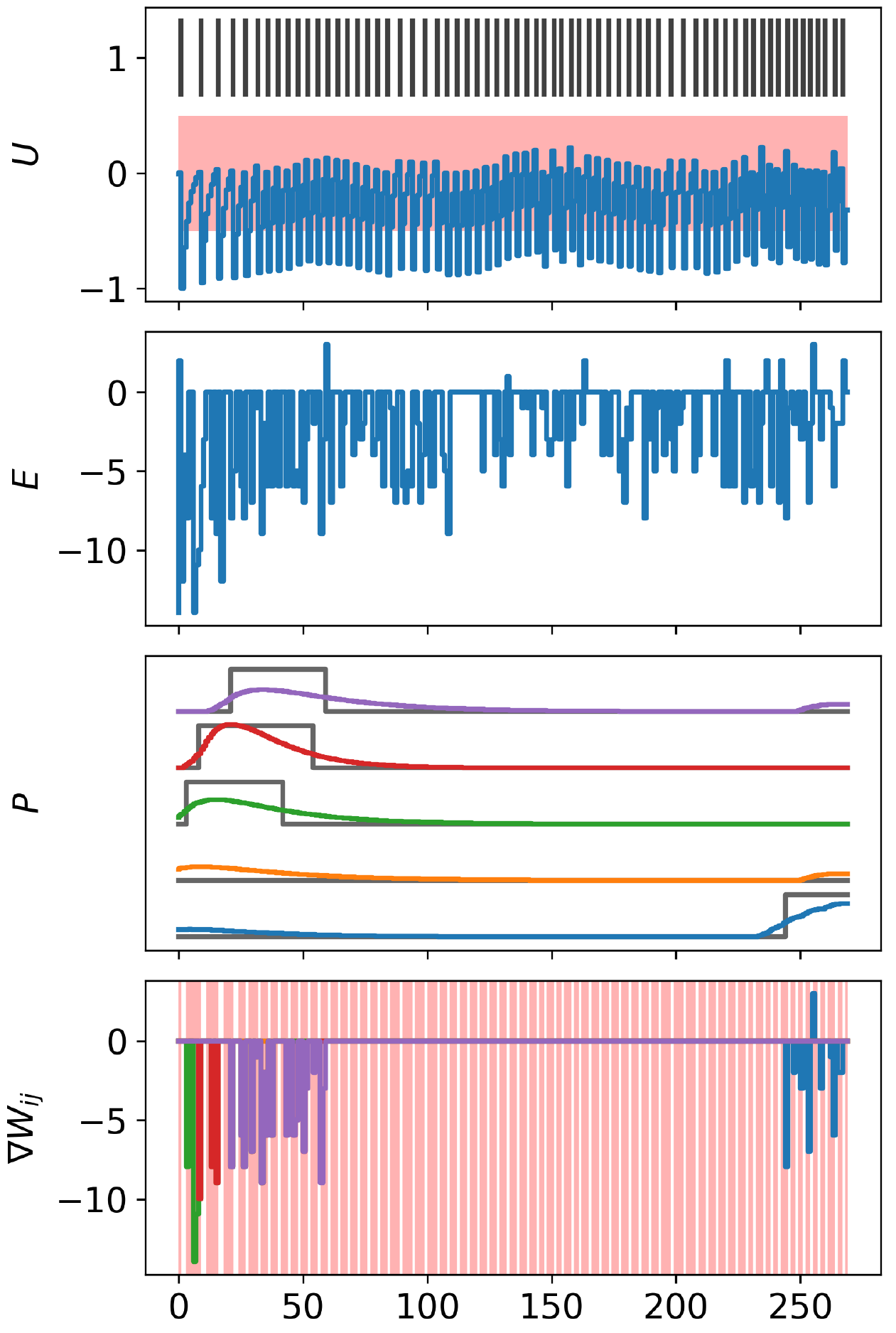}
    \includegraphics[width=.3\textwidth]{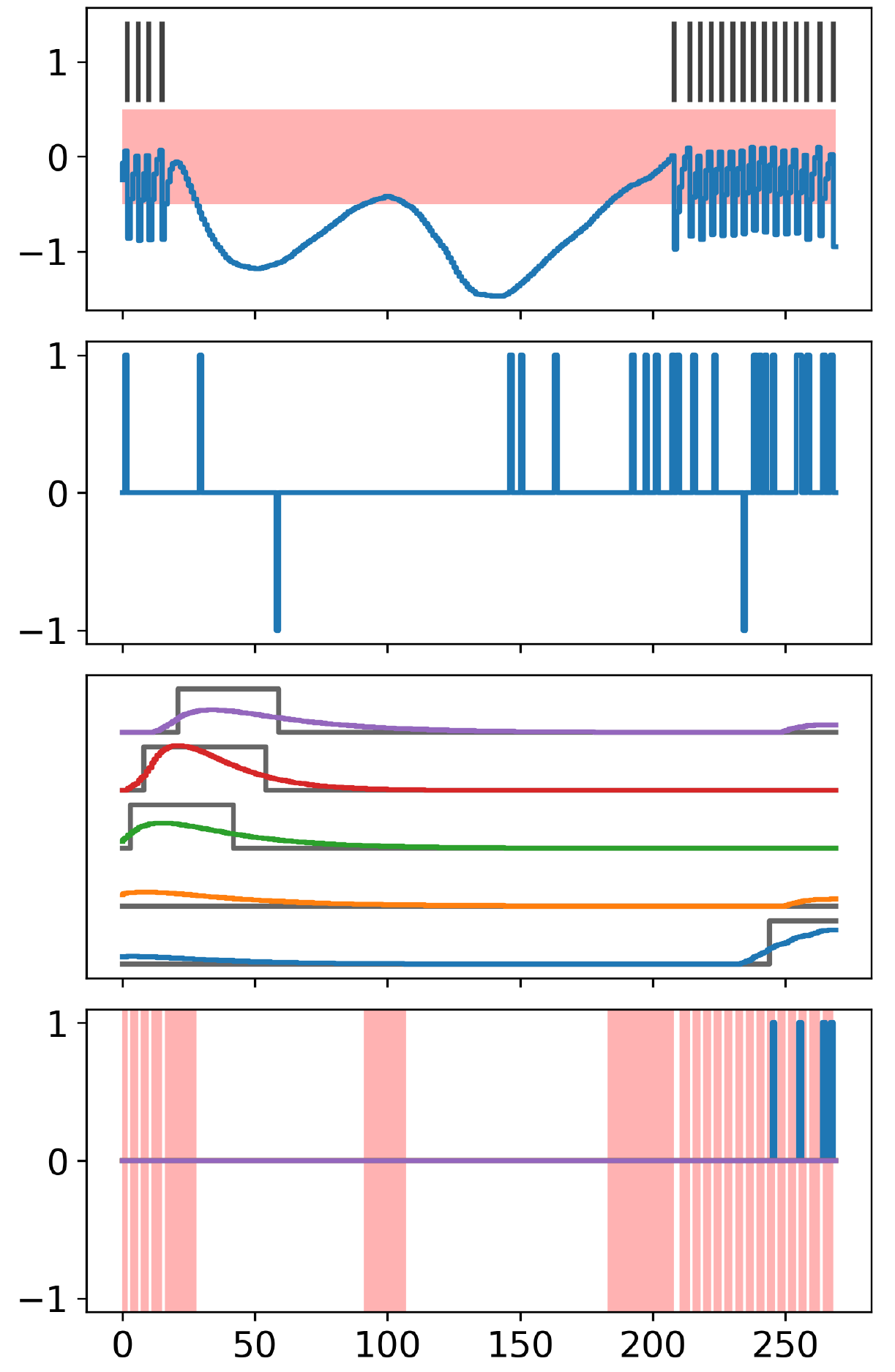}
    \includegraphics[width=.3\textwidth]{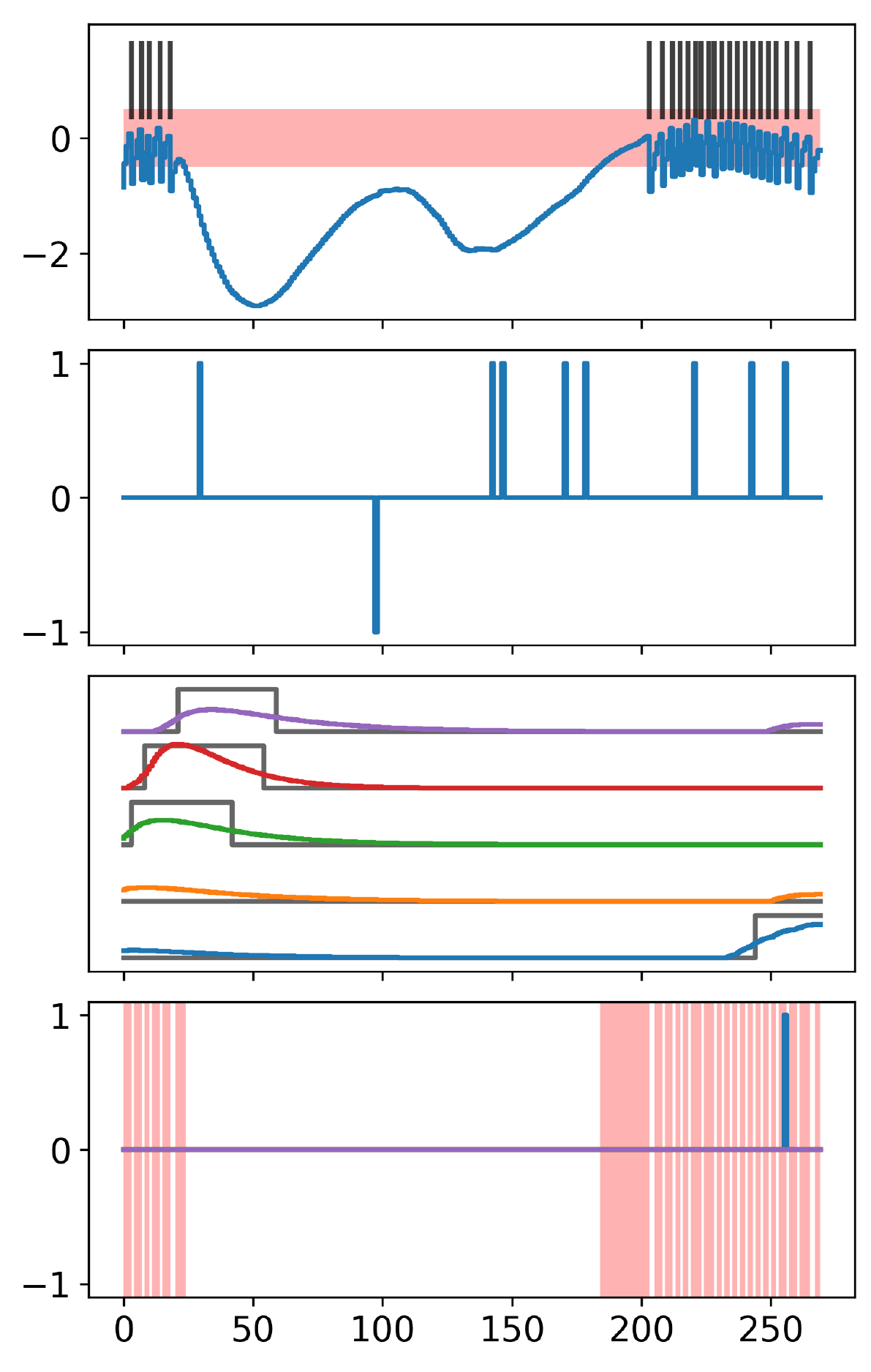}
    \caption{Signals during the learning of one N-MNIST sample at the start (epoch=0), middle (epoch=2) and end of learning (epoch=15) for a fully connected network. (Top) Membrane potential $U_i$ of neuron $i$ (blue) in layer $1$, overlaid with output spikes $S_i$ in the first layer. The red shading shows the region where $B_i=1$, \emph{e.g.} the neuron is eligible for an update. The fast, downwards excursions of the membrane potential are due to the refractory effect. (Middle-Top) Error events $E_i^1$ for neuron $i$. (Middle-Bottom) post-synaptic potentials $P_j$ for five representative synapses. The box-shaped curves show $\tilde{P}$ terms used to compute synaptic weight gradients $\nabla_{W_ij}$ for the shown synapses. (Bottom) Resulting weight gradients for the shown synapses. The red shading show regions where $B_i=1$. In these regions, if an error was present ($E_i\ne 0$), and $\tilde{P}>0$, then an update was made. Intuitively, learning corresponds to ``masking'' the values $E$ according to the neuron and synapse states.
    \label{fig:err_func_illustration}}
\end{figure*}
\begin{table*}[]
\caption{Recognition Error error-triggered learning over 3 runs.}
\centering
\begin{tabular}{c|c|c|c|c|c|c|c|c|}
\cline{2-9}& \multicolumn{4}{|c|}{\bf DVSGesture (ConvNet)} & \multicolumn{4}{c|}{\bf N-MNIST (MLP)} \\ 
\cline{2-9}& \multicolumn{2}{|c|}{$P$ }& \multicolumn{2}{c|}{\bf $\tilde{P}$}& \multicolumn{2}{c|}{\bf $P$} & \multicolumn{2}{c|}{\bf $\tilde{P}$}  \\ \hline

\multicolumn{1}{|c|}{$\bar{E}$} & Test Error        &  $\langle |E| \rangle$ & Test Error        & $\langle |E| \rangle$ & Test Error & $\langle |E| \rangle$ &Test Error & $\langle |E| \rangle$  \\ \hline
\multicolumn{1}{|c|}{1000Hz}& 4.07\% & $8.0 \cdot 10^6$ &  5.36\% & $8.0\cdot 10^6$ & 1.74\% &  $1.3 \cdot 10^6$  & 3.00 \%      & $1.3 \cdot 10^6$ \\
\multicolumn{1}{|c|}{50Hz}  & 5.62\% & $424 \cdot 10^3$&  7.12\% & $408 \cdot 10^3$& 2.34\% &  $66.8 \cdot 10^3$   & 4.52 \%     & $66.6\cdot 10^3$ \\
\multicolumn{1}{|c|}{10Hz}  & 6.52\% & $128 \cdot 10^3$ &  9.58\% & $96.7 \cdot 10^3$ & 3.51\% &  $14.5\cdot 10^3$  & 5.58 \%      & $14.7\cdot 10^3$  \\\hline
\end{tabular}
\label{tab:res}
\vspace{-0.15in}
\end{table*}

\section{Large Scale Experiments}\label{sec:sim_res}
An important feature of the error-triggered learning rule is its scalability to multi-layer networks with small and graceful loss of performance compared to standard deep learning.
To demonstrate this experimentally, we simulate the learning dynamics for classification in large-scale, multi-layer spiking networks on a GPU. 
The GPU simulations focus on event-based datasets acquired using a neuromorphic sensor, namely the N-MNIST and DVS Gestures dataset for demonstrating the learning model. 
Both datasets were pre-processed as in \cite{Kaiser_etal20_synaplas}.
The N-MNIST network is fully connected (1000--1000--1000), while the DVS Gestures network is convolutional (64c7-128c7-128c7).
In our simulations, all computations, parameters and states are computed and stored using full precision. However, according to the error-triggered learning rule, errors are quantized and encoded into a spike count.
Note that in the case of box-shaped synaptic traces, and up to a global learning rate factor $\tilde{\eta}$, weight updates are ternary (-1,0,1) and can in principle, be stored efficiently using a fixed point format.
For practical reasons, the neural networks were trained in mini-batches of 72 (DVS Gestures) and 100 (N-MNIST).
We note that the choice of using mini-batches is advantageous when using GPUs to simulate the dynamics and is not specific to \refeq{eq:spiking_neuron_rule}. 

Our model's parameters are similar to previous work \cite{Kaiser_etal20_synaplas}, except that the time constants were randomized.

The error rate, denoted $|E[t]|/1000$, is the number of non-zero values for $E[t]$ during one second of simulated time. The rate can be controlled using the parameter $\theta$. 
While several policies can be explored for controlling $\theta$ and thus $|E[t]|$, our experiments used a proportional controller with set point $\bar{E}$ to adjust $\theta$ such as the error rate per simulated second during one batch, denoted $\langle |E[t]| \rangle$, remains near $\bar{E}$.
After every batch, $\theta$ was adjusted as follows:
\[
\theta[t+1] = \theta[t] + \sigma (\bar{E} - \langle |E[t]| \rangle).
\]
where $\sigma$ is the controller constant and is set to $5\times10^{-7}$ in our experiments. Thus, the proportional controller increases the value of $\theta$ when the error rate is too large, and \emph{vice versa}. 

The results shown in \reftab{tab:res} demonstrate a small loss in accuracy across the two tasks when updates are error-triggered using $\bar{E} = .05$, and a more significant loss when using $\bar{E} = .01$. 
Published work on DVS Gestures with spiking neurons trained with backpropagation achieved 5.41\% \cite{Amir_etal17_lowpowe}, 6.36\% \cite{Shrestha_Orchard18_slayspik}, and 4.46\% \cite{Kaiser_etal20_synaplas} error rates and 1.3\% \cite{Lee_etal16_traideep} for N-MNIST with fully connected networks.
We emphasize here that the N-MNIST results are obtained using a multi-layer perceptron as opposed to a convolutional neural network. 
Spiking convolutional neural networks are capable of achieving lower errors on N-MNIST \cite{Esser_etal16_convnetw,Kaiser_etal20_synaplas}.

The results show final errors in the case of exact and approximate computations of $\pd{U}{W}$. Using the approximation $\tilde{P}$ instead of ${\pd{U}{W}}$ incurs an increase in error in all cases. This is because the gradients become biased. 
Several approaches could be pursued to reduce this loss: 1) using stochastic computing and 2) multi-level discretization of $\pd{U}{W}$.
A third conceivable option is to change the definition of $P$ in the neural dynamics such that it is also thresholded, so as to match $\tilde{P}$. 
However, this approach yielded poor results because $P_j$ became insensitive to the inputs beyond the last spike.

\reffig{fig:err_func_illustration} illustrates the signals used to compute $\Delta W_{ij}^l$ in the case of one N-MNIST data sample, at the beginning, middle and end of learning. 
There are many updates to the synaptic weights at the beginning of the learning, and several steps where $|err|>1$.
However, the number of updates regress quickly after a few epochs. The initial surge of updates is due to 1) a large error in early learning and 2) a suboptimal choice of $\theta_0$, the initial value of $\theta$. 
The latter could be optimized for each dataset to mitigate the initial surge of updates.

\begin{figure}
    \centering
    \includegraphics[width=.5\textwidth]{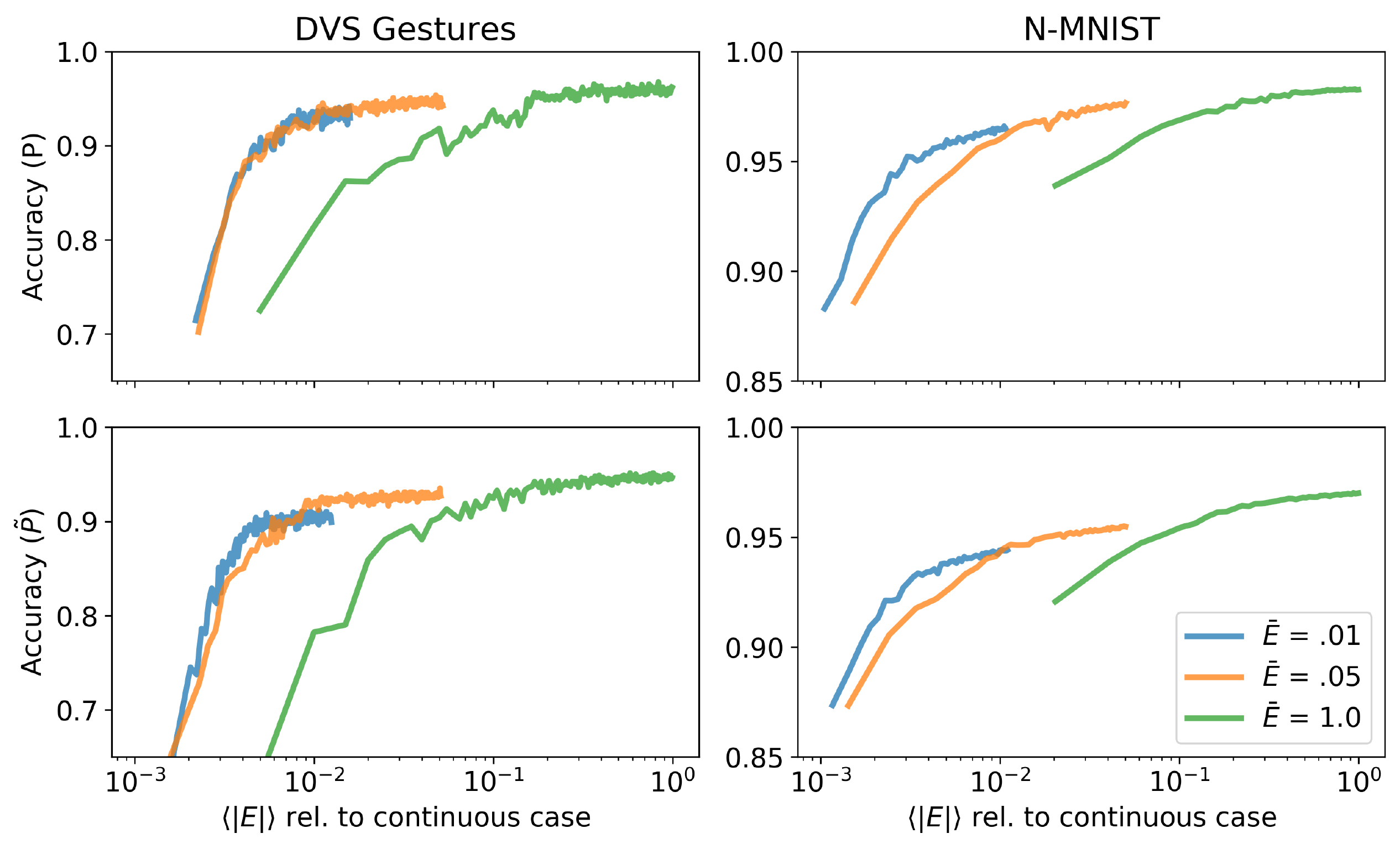}
    \caption{Accuracy -- error event-rate trade-off. The achieved accuracy as a function of the number of error events for DVS Gestures and N-MNIST datasets. The first row shows the results using the exact PSP, \emph{i.e.} $P$. The second row shows the results when using the approximate PSP $\tilde{P}$, respectively. For each experiment, three different target error rates $\bar{E}$ were selected. The horizontal axis show the total number of updates relative to the non-error-triggered case ($\bar{E}=1$). In all cases, $\bar{E}=.05$ provided nearly an order of magnitude fewer updates for a small cost in accuracy. 
    }
    \label{fig:error_vs_Rate}
\vspace{-0.8cm}
\end{figure}

It is conceivable that the role of event-triggered learning is merely to slow down learning compared to the continuous case. 
To demonstrate that this is not the case, we show task accuracy vs. the number of updates $\langle |E| \rangle$ relative to the continuously learning case in \reffig{fig:error_vs_Rate}.  
These curves indicate that values of $\bar{E}<1$ indeed reduce the number of parameters updates to reach a given accuracy on the task compared to the continuous case. 
Even the case $\bar{E} = .05$ leads to a drastic reduction in the number of updates with a reasonably small loss in accuracy.
However, a too low error event rate, here $\bar{E}=.01$ can result in poorer learning compared to $\bar{E} = .05$ along both axes (\emph{e.g.} \reffig{fig:error_vs_Rate}, bottom right, $\bar{E}=.01$). This is especially the case when the approximate traces $\tilde{P}$ are used during learning.
This implies the existence of an optimal trade-off for $\bar{E}$ that maximizes accuracy vs. the error rate. 

The weight updates are achieved through stochastic gradient descent (SGD). 
We note that SGD refers to how gradients are applied to the parameters, which is different from how the gradients are estimated. In this sense, SGD is consistent with the fact that errors are not backpropagated throught the multiple layers of the network. 
We used here SGD because other optimizers with adaptive learning rates (such as ADAM) with momentum involve further computations and states that would incur an additional overhead in a hardware implementation.
To take advantage of the GPU parallelization, batch sizes were set to 72 (DVS Gestures) and 200 (N-MNIST). 
Although, batch sizes larger than 1 are not possible locally on a physical substrate, training with batch size 1 is just as effective as using batches \cite{LeCun_Bottou04_largscal}. 
Our earlier work demonstrated that training with batch size 1 in \acp{SNN} is indeed effective \cite{Neftci_etal17_evenranda, Detorakis_etal18_neursyna}, but cannot take advantage of GPU accelerations.

\begin{figure*}[t]
  \centering\includegraphics[width=\textwidth]{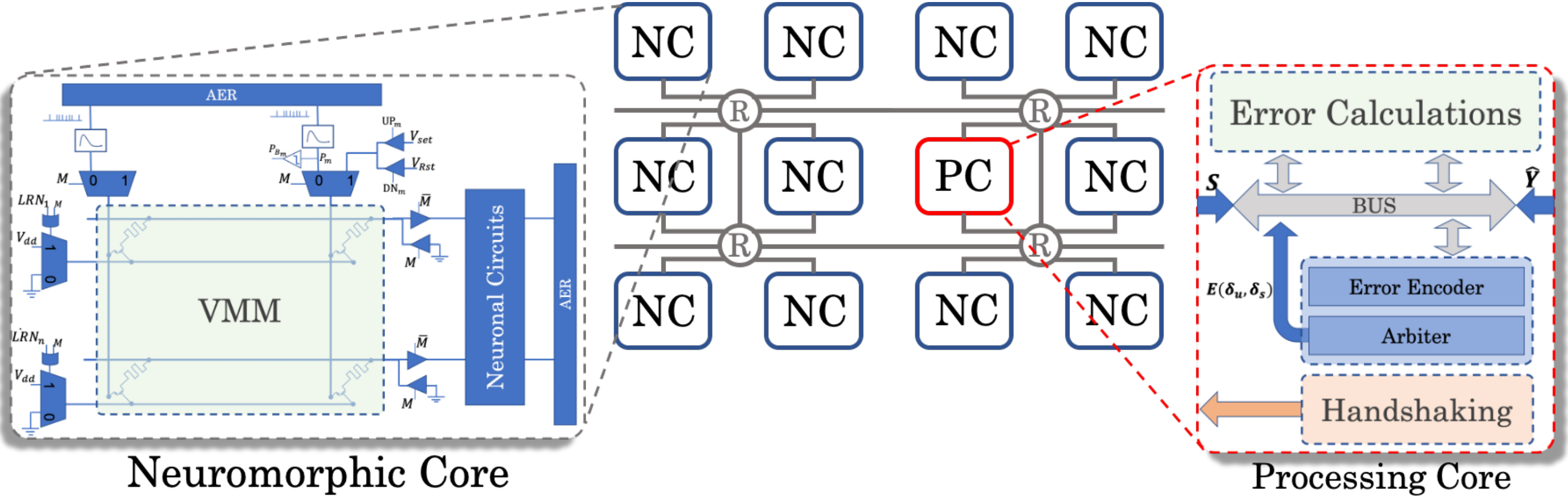}
  \caption{Details of the hardware architecture consisting mainly from neuromorphic and processing cores.}

  \label{fig:archOverview}
 \vspace{-0.5cm}
\end{figure*}

\section{Hardware Architecture}\label{sec:hw_arch}

In this section, we discuss the general hardware architecture that supports the error-triggered learning, which comprises \acp{NC} and \acp{PC} as depicted in Fig. \ref{fig:archOverview}.
The \acp{NC} are responsible for implementing the neuron, synapse dynamics \refeq{eq:lif_equations}, and the estimation of $\pd{S^{l}}{W^{l}}$.
Each core additionally contains circuits that are needed for implementing training provided the error signals. 
The error signals are calculated on the \acp{PC} and communicated asynchronously to the \acp{NC}. 

The separation in \ac{PC} and \ac{NC} is directly motivated by the separation of \refeq{eq:binary_neuron_rule} in extrinsic and intrinsic factors as follows:
\[ 
 \Delta W_{ij}^l \propto \underbrace{E^l_i}_{PC} \underbrace{B(U_i^l) P_j^{l}}_{NC}. 
\]
This separation results in an elegant, and largely local estimation of the gradients since $P_j^l$ above is already available in the neural dynamics, and $B(U_i^l)$ is simply a thresholded copy of $U_i^l$. 
Thus, the \ac{NC} contributes to inference and gradient estimation, whereas the estimation of the error $E^l_i$ is carried out in the PC and communicated to the NC.
The ternary nature of the error further reduces the overhead of communicating these errors across the two types of cores.

\subsection{Processing Core}\label{sec:error_core}
In addition to data and control buses, the \ac{PC} consists of four main blocks, namely for error calculation, error encoding, arbitration, and handshaking. 
The \ac{PC} can be shared among several \acp{NC}, where communication across the two types of cores is mediated using the same address event routing conventions as the \acp{NC}. 

The error calculation block is responsible for calculating the gradients and the continuous-value of the error updates (\emph{i.e.}, $err^l$ signals).  
The PC also compares the error signal $err$ with the threshold $\theta$ 

as discussed in \eqref{eq:error-coding-neurons} and \eqref{eq:binary_neuron_rule} to generate integer $E$ signals that are sent to error encoder. 
A natural approach to implement this block is by using general-purpose Central Processing Unit (CPU) in addition to a shared memory which is similar to the Lakemont processors on the Intel Loihi research processor \cite{Davies_etal18_loihneur}. 
CPUs offer high speed, high flexibility, and programming ability that is generally desirable when calculating loss functions and their gradients. 
The shared memory can be used to store the spike events while calculating a different layer error. 

The calculated error update signals $E$ are rate-encoded in the error encoder into two spike trains $E\rightarrow{\{\bf{\delta}_u, \bf{\delta}_s\}}$ where $\bf{\delta}_u$ is the update signal and $\bf{\delta}_s$ is the polarity of the update. 

The arbiter is used to choose only one \ac{NC} to update at time. This choice can be based on different policies, for instance, least frequently updated or equal policy. 
Once the $\{\bf{\delta}_u, \bf{\delta}_s\}$ signals are generated, they need to be communicated to the corresponding \ac{NC}.

For this communication, a handshaking block is required. The generated error events send a request to the \ac{PC} arbiter, which acknowledges one of them (usually based on the arrival times). The address of the acknolwedged event along with a request is communicated to the \ac{NC} core in a packet. The handshaking block at the \ac{NC} ensures that the row whose address matches the packet receives the event and takes control over the array. This block then sends back an acknowledge to the \ac{PC} as soon as the learning is over. The communication bus is then freed up and is made available for the next events.

An alternative to implementing the \ac{PC} is to use another \ac{NC}, as it is a \ac{SNN} that can be naturally configured to implement the necessary blocks for communication and error encoding. 
General-purpose functions can be computed in \acp{SNN}, for example, by using the neural engineering framework \cite{Eliasmith_Anderson04_neurengi}. 
In this case, the system could consist solely of \acp{NC}.
The homogeneity afforded by this alternative may prove desirable for specific technologies and designs.

\vspace{-0.4cm}
\subsection{Neuromorphic Core}\label{sec:Xbar}
Emerging technologies, such as Resistive RAM (RRAMs), Phase Change Memories (PCMs), Spin Transfer Torque RAMs (STT-RAMs), and other MOS realizations such as floating gate transistors, assembled as an \ac{RCA} enable the \ac{VMM} operation to be completed in a single step \cite{sebastian2020_review}. This is unlike general-purpose processors that require $N\times M $ steps where $ N $ and $ M $ are the weight matrix's size.
Recently, there has been impressive development in maturing the technology and large array sizes for PCM and OxRAMs have already been reported \cite{sebastian_etal2017_million, azzaz_etal2016_16kb}.
These emerging technologies implement only positive weight (excitatory connections). However, to fully represent the neural computations, negative weights (inhibitory connections) are also necessary.
There are two ways to realize the positive and negative weights ~\cite{ Fouda_etal18_indecomp}. 1) Balanced realization where two devices are needed to implement the weight value stored in the devices conductances where $W=G^+-G^-$.
If the $G^+$ is greater/less than $G^-$, it represents positive/negative weight, respectively.
2) Unbalanced realization where one device is used to implement the weight value with a common reference conductance $G_{ref}$, set to the mid-value of the conductance range.
Thus, the weight value is represented as $W=G-G_{ref}$. 
If the $G$ is greater/less than $G_{ref}$, it represents a positive/negative weight, respectively.
In this work, we use an unbalanced realization since it saves area and power at the expense of using half of the device's dynamic range. 
Thus, the memristive SNN can be written as:  
\begin{equation}\label{eq:lif_memristor_equations}
    U_i^l[t] = \sum_j \left(G^{l}_{ij}-G_{ref}\right)P_j^l[t] - \delta R_i [t].
\end{equation}
By following the same analysis in section III-A, the dynamics are the same as \refeq{eq:spiking_neuron_rule}. 

\acp{NC} implement the presynaptic potential circuits that simulate the temporal dynamics of $P$ in \refeq{eq:lif_equations}.
In addition, the \ac{NC} implements the memristor write circuitry which potentiate or depress the memristor with a sequence of pulses depending on the error signal that is calculated in the \ac{PC}. 
The \ac{NC} continuously works in the inference mode until it enters the learning mode by receiving an error event from the \ac{PC}. 

The circuit then deactivates all rows except the row where the error event belongs to. 

The memristors within this row are then updated by a positive or negative pulse based on the $\tilde{P}$ value, which would potentiate or depress the device by $\pm \Delta G$ as shown in \reftab{Update_table}. 
Thus, the control signals can be written as follows:
\[\text{UP}_j =\bar{\delta_s} \tilde{P}_j \quad \text{DN}_j =\delta_s \tilde{P}_j \]
\[lrn_i =B_i \delta_u\quad LRN  =\sum_i lrn_i \]
where $LRN$ is the mode signal which determine the mode of the operation either inference ($LRN=0$) or weight update mode ($LRN=1$). The update mode is chosen if any of the $lrn$ signals is turned {\verb ON }. The full details of the neuromorphic core is discussed in \refsec{sec:neurocore}.

It is worth to mention that we considered local learning where each layer learns individually. As a result, there is no backpropogation as known in the conventional sense. The loss gradient calculations are performed in the processing core with floating point precision to calculate the error signals. These are then quantized and serially encoded into ternary pulse stream to program the memrsitors. Strategies for reduced precision learning (e.g. stochastic rounding) are compatible with error triggered learning \cite{gupta2015deep}.

\vspace{-0.2cm}
\subsection{Network On Chip}
The neuromorphic and processing cores are linked together with a Network on Chip (NoC) that organizes the communication among them based on the widely used \ac{AER} scheme. \cite{deiss1999pulse,Lazzaro_etal93_siliaudi}. 
Different routing techniques have been proposed to trade off between flexibility (i.e., degree of configurablity) and expandability \cite{Park_etal17_hieraddr}. 
For instance, TrueNorth and Loihi chips use 2D mesh NoC \cite{merolla2014million,Davies_etal18_loihneur}, SpiNNiker uses torus NoC \cite{painkras2013spinnaker} and HiAER uses tree NoC \cite{Park_etal17_hieraddr}. 
HiAER offers high flexibility and expandability, which can be used in the proposed architecture for communication among neuromorphic cores during inference and between the processing core and neuromorphic cores during training. 

\begin{table}[!t]
\centering
\caption{Truth table of the error-triggered ternary update rule. Signals $B$, and  $\tilde{P}$ are local to the \ac{NC}, while signals $E$, $LRN$, $UP$ and $DN$ are local to the \ac{PC}.}
\begin{tabular}{|c|c|c|c|c|c|c|c|}
\hline
\multicolumn{2}{|c|}{$\eta E$}& \multicolumn{1}{c|}{\multirow{2}{*}{$B$}} & \multicolumn{1}{c|}{\multirow{2}{*}{$\tilde{P}$}} & \multicolumn{1}{c|}{\multirow{2}{*}{$W$}} & \multicolumn{1}{c|}{\multirow{2}{*}{$LRN$}} & \multicolumn{1}{c|}{\multirow{2}{*}{UP}} & \multicolumn{1}{c|}{\multirow{2}{*}{DN}} \\ \cline{1-2}
\multicolumn{1}{|c|}{$\delta_u$} & \multicolumn{1}{c|}{$\delta_s$} & \multicolumn{1}{c|}{}  & \multicolumn{1}{c|}{}  & \multicolumn{1}{c|}{}  & \multicolumn{1}{c|}{} & \multicolumn{1}{c|}{}   & \multicolumn{1}{c|}{} \\ \hline 
0 & $\times$ & $\times$   & $\times$    & 0 & 0   & 0  & 0  \\ \hline
1   & $\times$  & 0 & $\times$  & 0  & 0  & 0  & 0  \\ \hline
1   & $\times$  & $\times$ & 0  & 0 & 0 & 0 & 0     \\ \hline
1   & 0  & 1 & 1 & $+\Delta G$ & 1 & 1  & 0       \\ \hline
1  & 1 & 1  & 1 & $-\Delta G$ & 1 & 0   & 1       \\ \hline
\end{tabular}
\label{Update_table}
\end{table}

\subsection{Hardware Limits}
\label{sec:hwlim}

\paragraph*{Error frequency} A full update cycle of the \ac{NC} is $T_{u_{max}}= N\times f_{er_{max}}\times T_p$ where $N$ is the fan-out per \ac{NC}, $f_{er_{max}}$ is the maximum error frequency and $T_p$ is the width of the memristor update period. $T_{u_{max}}$ should be much smaller than the inter-spike interval (\emph{i.e.} factor of 10 will be sufficient).
Assuming that the maximum firing rate of the neuron is $f_{n_{max}}$, a condition on maximum error frequency can be derived as
\[
f_{er_{max}}<\frac{1}{10N f_{n_{max}} T_p}
\]
This shows a trade off between the fan-out per \ac{NC} and the maximum error frequency. If we considered $T_p=100\,$ns \cite{ielmini2020_review} and $f_{n_{max}}=100\,Hz$, the maximum error frequency under this definition is $78$\,Hz for $N=128$ (a typical size of the current fabricated \ac{RCA}) and $10$\,Hz for $N=10^3$. As previously evaluated in  \refsec{sec:sim_res}, the higher the error frequency, the better the performance. The hardware would set the upper limit for the error frequency to $10$ Hz for $N=10^3$, which causes 2.68\% and 4.22\% drop in the performance. Depending on the distribution of the spike trains from the \emph{error calculation} block, this constraint can be further loosened. 
While a buffer can also be added to the \ac{PC} to queue the error events which are blocked as a result of the busy communication bus, this translates to more memory and hence area on the \ac{PC} and lead to biased gradients.

\paragraph*{Input frequency}
A similar analysis can be done to calculate the maximum input dimension of the array. Assuming there is no structure in the incoming input (or that the structure is not available \emph{a priori}), a Possion statistic can be considered for the input spikes. In that case, the probability of the next spike in any of the $M$ inputs occurring within the pulse width of the write pulse $T_p$ is equal to $P(Event)=1-e^{-M f_{in}T_p}$ where $f_{in}$ is the frequency of the input spikes. 
To keep this probability low (e.g., $<0.01$), the fan-in can be calculated. 
Considering a biologically plausible maximum rate of $f_{in} = 100$\,Hz, in the worst case where all input neurons fire and for $T_p=100\,$ns, the maximum $M$ would be 1000. The \ac{SNN} test benches, such as DVSGesture and N-MINST that have been discussed in \ref{sec:surgrad}, have peak event rates around 30\, Hz and 15\,Hz  \cite{Lee_etal16_traideep} repectively which would triple the fan-in of the \ac{NC}.  

\paragraph*{Number of \ac{NC}s per \ac{PC}} 
Assuming that the PC runs at frequency $f_{clk}$, and it takes $2 N/f_{clk}$ on average to calculate the error signals (which can be $2/f_{clk}$ in the case of a \ac{RCA} or $2N^2/f_{clk}$ in case of a von-Neumann architecture). 
The factor $2$ is added for $J$ and $H$ multiplications in addition to loss calculation evaluation time $T_l$. 
Thus, the total error calculation per NC takes $T_{pc}=2 N/f_{clk}+T_l$. 
Updates have to be performed faster than the time constant for computing the gradient. Thus, the maximum number of NCs is $N=T_{pc}/T_{u_{max}}$. 
For example, for $f_{clk}=500$\,MHz and $N=1000$ and $T_{u_{max}}=1$\,ms, $4000$ \acp{NC} can be used per \ac{RCA}-based \ac{PC} on average and $4$ \acp{NC} for von-Neumann-based \ac{PC}. It is worth noting that handshaking, arbiter and the error encoder are operating in parallel with the error calculations and thus we did not include them in the estimation.

\section{Neuromorphic Core Implementation}
\label{sec:neurocore}

\begin{figure*}[t]
  \centering\includegraphics[width=0.95\textwidth]{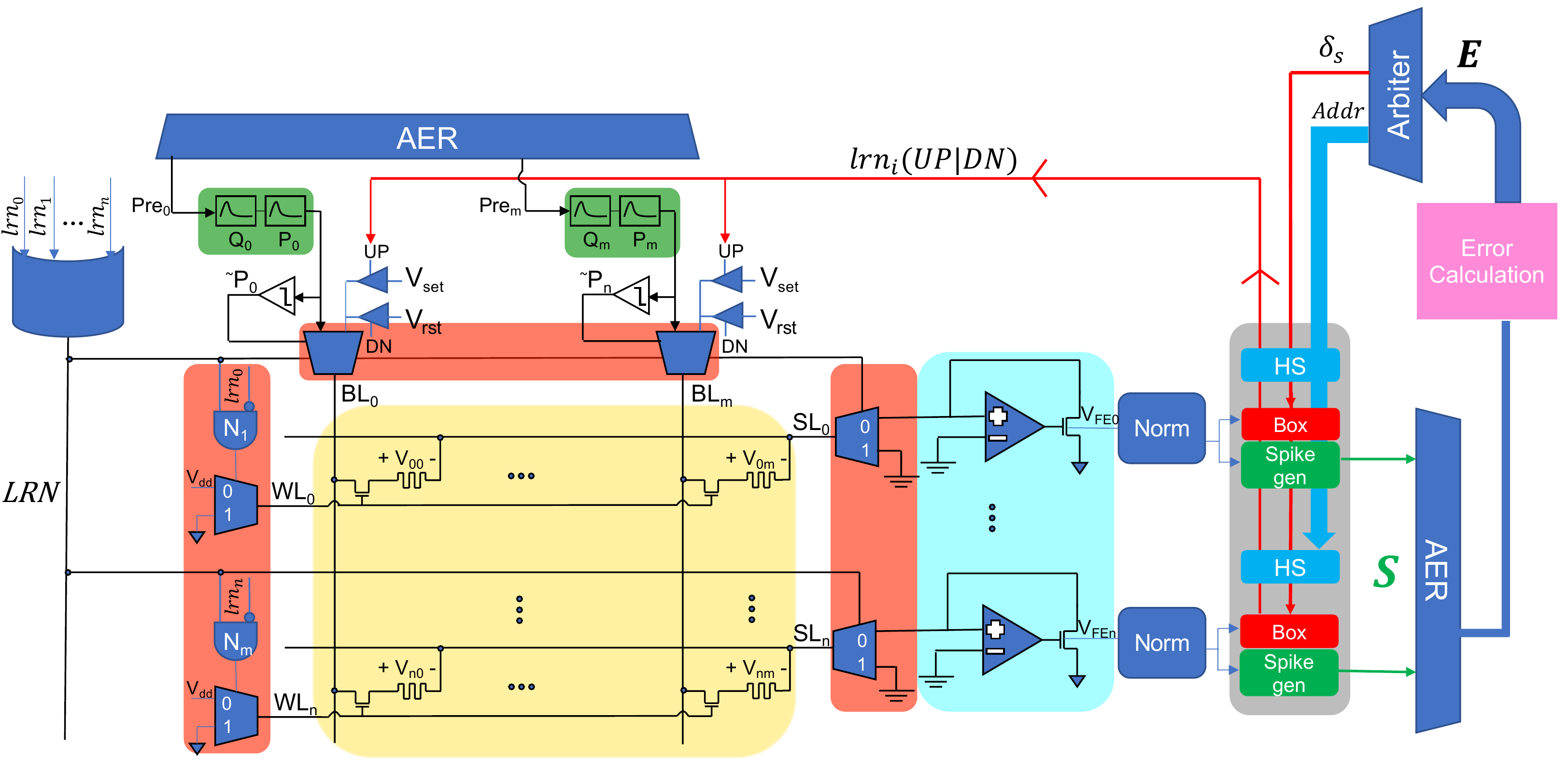}
  \caption{Details of the neuromorphic learning architecture compatible with the \ac{AER} communication scheme. {The event integrators, VMM array, switches, front end, neural circuits, and Error calculation block are highlighted in green, yellow, orange, light blue, gray and pink respectively. The color code matches the high-level architecture in Fig.~\ref{fig:Idealarch}}. The 1T-1R array of memristive devices is driven by the appropriate voltages on the \ac{WL}, \ac{SL}, \ac{BL} for inference and learning. During inference the voltages across the memristor are proportional to the respective $P$ value. The current from the \ac{RCA} gets normalized in the \emph{Norm} block which is fed to the \emph{box} and \emph{spike gen} blocks. The spikes $S$ from the \emph{Spike gen} are given to the \emph{error calculation} block which sends the arbitrated error events with the address of the learning row to the handshaking blocks (\emph{HS}). This communication gives the control of the array to the learning row which sends back the $lrn_i$ signals back to the \ac{RCA}.}
  \label{fig:arch}
    \vspace{-0.15in}
\end{figure*}

\begin{figure}[t]
  \centering\includegraphics[width=0.35\textwidth]{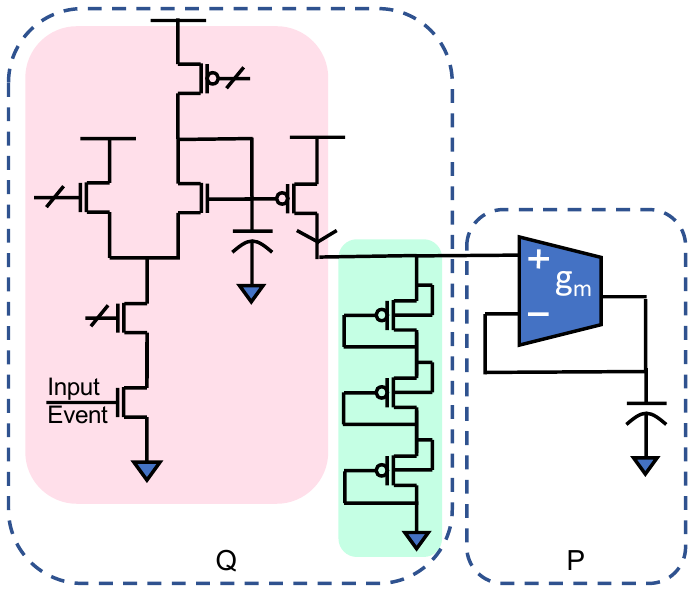}
  
  \caption{Double integration scheme using Q and P integrators as is explained in Fig.~\ref{fig:archOverview}. Q integrator is a \ac{DPI} circuit in pink whose output current is converted to a voltage by the pseudo resistor highlighted in green. P integrator is a $G_m C$ filter.
  }
  \label{fig:PQ}
    \vspace{-0.15in}
\end{figure}

\begin{figure*}[t]
  \centering\includegraphics[width=0.8\textwidth]{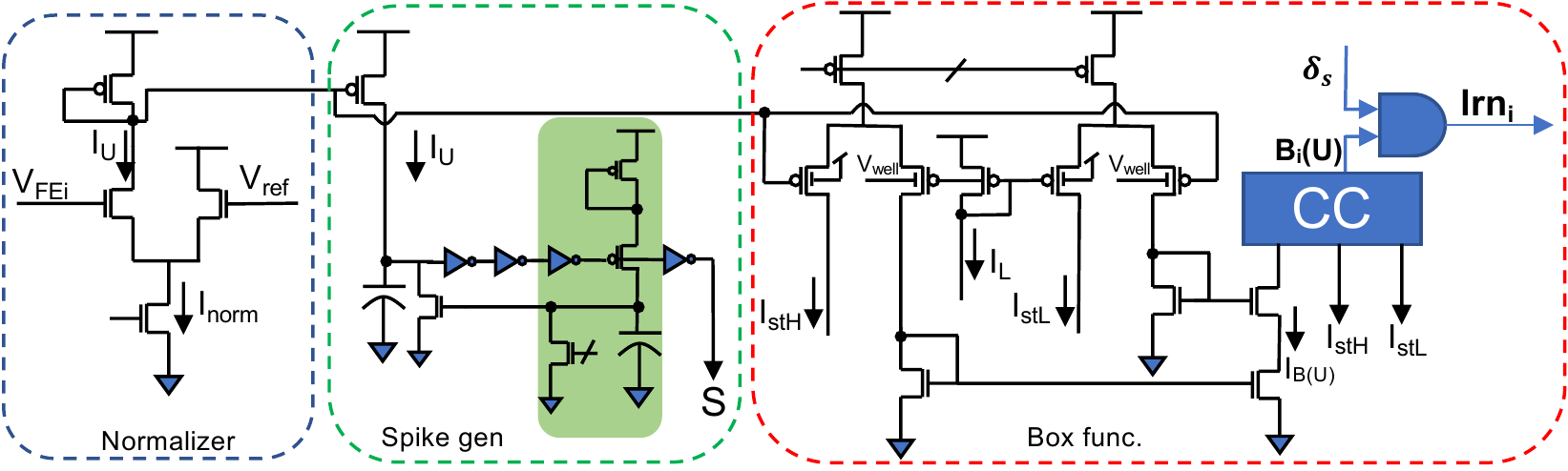}
  \caption{The learning circuits representing normalizing, spike generation and box functions. The normalizer circuit normalizes the current from the crossbar array to $I_{norm}$ set by the tail of the differential pair. Spike generator circuit is a simple current to frequency (C2F) converter generating spike $S$ of the neurons. The highlighted part depicts the circuit implementing the refractory period to limit the firing rate and hence power consumption of the block. The box function gates the learning signals $UP$ and $DN$ by its output $B(U)$. It is implemented with a bump circuit where the bump current $I_{B(U)}$ gets compared to the anti bump currents by current comparator (CC) and when higher, $B(U)$ has a binary value of 1 allowing learning to happen. 
 }
  \label{fig:neuralcircuits}
\vspace{-0.2cm}
\end{figure*}

In this section, we will first introduce the neuromorphic learning architecture compatible with a $1T-1R$ \ac{RCA}, and the signal flow from the input events to the learning core. We will then detail the circuits that implement the building blocks of the architecture. 

\subsection{Learning Architecture}
\paragraph{Overall information flow} Our \ac{SNN} circuit implementation differs from classical ones used in mixed-signal neuromorphic chips \cite{Indiveri_Chicca11_vlsineur}.
Generally, the rows of crossbar arrays are driven by spikes and integration takes place at each column \cite{Chen_etal15_mitieffe,qiao_etal2015_rolls,Payvand_etal19_neursyst,Park_etal14_65k-73-m}. 
While this is beneficial in reducing read power, it renders learning more difficult because the variables necessary for learning in \acp{SNN} are not local to the crossbar. 
Instead, we use the crossbar as a \ac{VMM} of pre-synaptic trace vectors $P^l$ and synaptic weight matrices $W^l$. 
Using this strategy, the same trace $P_i^l$ per neuron supports both inference and learning. 
This property has the distinctive advantage for learning in that it is immune to the mismatch in $P^l_i$, and can even exploit this variation. \ac{AER} is the conventional scheme for communication between neuronal cores in many neuromorphic chips \cite{Deiss_etal98_pulscomm}. \reffig{fig:arch} depicts the details of the neuromorphic learning architecture as a crossbar compatible with the \ac{AER}. {The information flows from the \ac{AER} at the input columns to the integrators, then to the VMM and finally to the spike generator block which sends the output spikes to the row \ac{AER}. Through the row \ac{AER}, information then flows to the \ac{PC} to calculate the error, which in turn sends error events back to the \ac{VMM} to change the synaptic weights. }

\paragraph{Details of the architecture }Pre-synaptic events communicated via \ac{AER} are integrated in the $Q$ blocks, which are then integrated in $P$ blocks as shown in \reffig{fig:arch}. 
This doubly integrated signal then drives the \ac{RCA} during inference mode. The \ac{RCA} model used here is a $1T-1R$ array of memristive devices with the gate and source of the transistor being driven by the \ac{WL} and \ac{BL} respectively and the bottom electrode of the device being driven by the \ac{SL}. 
The voltages driving the \ac{WL}, \ac{BL} and \ac{SL} are MUX-ed at the periphery to drive the array with the appropriate voltages depending on the inference or learning mode. {It is worth noting that in our simulations, we did not use a specific model for the devices. Any type of device whose conductance can be changed with a voltage pulse can be used in this type of architecture. Specifically, our architecture matches well with Oxide-based Resistive RAM (OxRAM) \cite{yu_etal2011_oxram} and Conductive Bridge RAM (CBRAM) type of devices \cite{suri_etal2013_cbram}.}

In inference mode, \ac{WL} is set to $V_{dd}$ which turns on the selector transistor, \ac{BL} is driven by buffered $P$ voltages, and the \ac{SL} is connected to a \ac{TIA} which pins each row of the array to a virtual ground. 
The current from the \ac{RCA} is dependent on the value of the memristive devices. 
To ensure subthreshold operation for the next state of the computation, a normalizer block is used.
The normalized output is fed both to a spike generator (spike gen) and a learning block (box). 
The pulse generator block acts as a neuron that only performs a thresholding and refractory function since its integration is carried out at the $P$ block. 
The generated $S$ spikes are communicated to the error generator block through the \ac{AER} scheme as well as other layers.
The learning block generates the box function described in \refeq{eq:spiking_neuron_rule}.

In the learning mode, the array will be driven by the appropriate programming voltages on \ac{WL}, \ac{BL} and \ac{SL} to update the conductance of the memristive devices. 
Since the whole array will be affected by the \ac{BL} and \ac{SL} voltages, at any point in time only one row of devices can be programmed.
Since in our approach, the updates will be done on the error events which are generated per neuron, this architecture maps naturally to the error-triggered algorithm as the error events are generated for each neuron and hence per row. 
The error events are generated through the \emph{error calculation} block shown in~\reffig{fig:arch}.
This block can be implemented by another \ac{SNN} or any non-linear function of $S$ implemented by a digital core (\emph{e.g.} as explained in \refsec{sec:error_core}). 
The calculated errors are encoded in {\verb UP } and {\verb DN } learning events for every neuron of the array.
Since only one neurons' synapses can be updated at any point in time, these learning signals are arbitrated and the access to the learning bus will be granted to the learning signals of one neuron.  As is explained in \refsec{sec:error_core}, the address of the acknowledged neuron  is sent to the Handshaking blocks (HS) at each row (through $Addr$ bus shown in~\reffig{fig:arch}) along with the sign of the update ($\delta_s$). The corresponding row $i$ whose address matches $Addr$ receives the address and its box block generates the $lrn_i$ signal depending on the $B$ value as specified in~\reftab{Update_table}.
The $\ac{WL}_i$ remains at $V_{dd}$ and all the other $\ac{WL}_j, j\neq i$ switch to zero such that neuron $i$ takes control over the array 
(implemented by gates $N_i$ in~\reffig{fig:arch} which perform the AND operation between $\tilde{lrn_i}$, and $LRN$ signals which is the output of the OR operation between all $lrn_i$ signals); 
Once in the learning mode indicated by the OR output ($LRN$ signal), \ac{SL} is switched to a common mode voltage (virtual ground) which blocks learning signals to the neurons. 
The voltage on $\ac{BL}_i$ (hence the $V_{ij}$ in the figure) depends on the state of $\tilde{P}_j$ which is a binary value as a result of comparing $P_j$ with a threshold as is shown in~\reffig{fig:arch}. 
In accordance with the truth table~\reftab{Update_table}, on the arrival of the { \verb UP } or { \verb DN } event, if $B_i$ and $\tilde{P}_j$ are non-zero, voltage $V_{set}$ or $V_{rst}$ will get applied to $BL_i$ respectively.

Once learning is over, the handshaking block elicits an acknowledge signal to the \emph{error calculating} block which frees up the array and the $Addr$ and $\delta_s$ wait for the next request.

\vspace{-0.2cm}
\subsection{Detail of the Neuromorphic Circuits}
At the arrival of the $Pre_j$ events from the \ac{AER} input, the trace $Q_j$ is generated through a \ac{DPI} circuit shown in pink in \reffig{fig:PQ} which generates a tunable exponential response in the form of a sub-threshold current \cite{Bartolozzi_Indiveri07_synadyna}. 
The current is linearly converted to voltage using pseudo resistors in the block highlighted in green in \reffig{fig:PQ}. 
The first-order integrated voltage is fed to a $G_m C$ filter giving rise to a second-order integrated output $P_j$ which is buffered to drive the entire crossbar column in accordance with \refeq{eq:lif_equations}.

Output voltage $P_j$ is applied to the top electrode of the corresponding memristive device ($W_{ji}$) whose bottom electrode is pinned by the crossbar front-end \ac{TIA}.
This block pins the entire row to virtual ground (in our case common mode voltage is set as half $V_{dd}$) and reads out the sum of the currents generated by the application of $P$s across the memristors in the row.
As a result, voltage $V_{FEi}$ is developed on the gate of the transistor at the output of the \ac{TIA} which feeds to a normalizer circuit shown in \reffig{fig:neuralcircuits}. This circuit is a differential pair which re-normalizes the sum of the currents from the crossbar to $I_{norm}$, ensuring that the currents remain in the sub-threshold regime for the next stage of the computation which is (i) the box function $B(U)$ as is specified in \refeq{eq:error-coding-neurons} implemented by the box block and (ii) the spike generation block which gives rise to $S$.\\ 
\noindent (i) is carried out by a modified version of the Bump circuit \cite{delbruck1991bump} which is a comparator and a current correlator that detects the similarity and dissimilarity between two currents in an analog fashion, as shown in the red box of Fig~\ref{fig:neuralcircuits}.  
The width of the transfer characteristics of the Bump current directly implements the box function $B(U)$ where $I_U$ is close to $I_L$ (condition $u_-<U_i^l<u_+$ from \refsec{sec:surgrad}).
Unlike the original Bump circuit, \ac{VWBump} 
\cite{minch2016simple} enables configurability over the width of the box function by changing the well potential $V_{well}$. Moreover, the tunability of $I_L$ allows setting the offset of the box function with respect to the normalized crossbar current. 
The details of using this circuit for learning is explained in~\cite{Payvand_Indiveri19_spikplas}.
The output of \ac{VWBump} circuit gates the arbitrated \verb UP  and  \verb DN  signals from the \ac{PC} to indicate the sign of the weight update (up or down) or stop-learning (no update).\\
\noindent (ii) is carried out via a simple current to frequency (C2F) circuit, which directly translates $I_U$ to spike frequency $S$. The highlighted part implements the refractory period, which limits the spiking rate of this block. 
\vspace{-0.2cm}
\subsection{Circuit Simulations Results}

In this section, we report the simulations results for a standard CMOS 180\,nm process, showing the characteristics and output of the learning blocks. Moreover, we present the voltages across the memristive devices in a learning loop. 

\begin{figure}[t]
  \centering\includegraphics[width=0.5\textwidth]{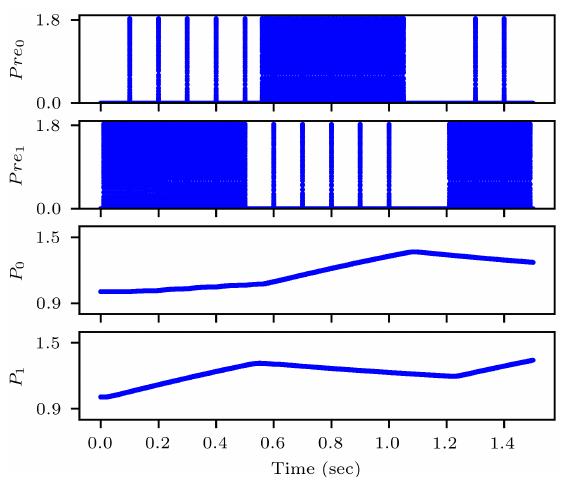}
      \vspace{-0.1in}
  \caption{Input events on two different input channels and their double integrated output $P_0$ and $P_1$. 
  }
  \label{fig:inputtrace}
    \vspace{-0.15in}
\end{figure}

\reffig{fig:inputtrace} shows the output of the double integration of the input events $Pre_i$ coming from the \ac{AER}. $Pre_0$ and $Pre_1$ and subsequently $P_0$ and $P_1$ are plotted as examples. $P_i$ smoothly follows the instantaneous firing rate of $Pre_i$ as is expected. 
\begin{figure}
  \begin{center}
    \begin{subfigure}[]{0.25\textwidth}
      \includegraphics[width=0.9\textwidth]{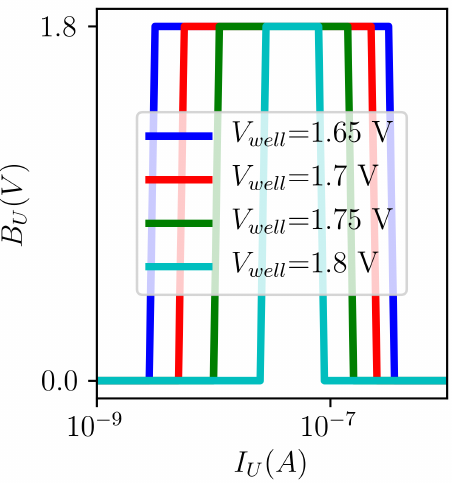}
      \caption{}
      \label{fig:B_vwell}
    \end{subfigure}
    \hspace{-0.5cm}
    \begin{subfigure}[]{0.25\textwidth}
      \includegraphics[width=0.95\textwidth]{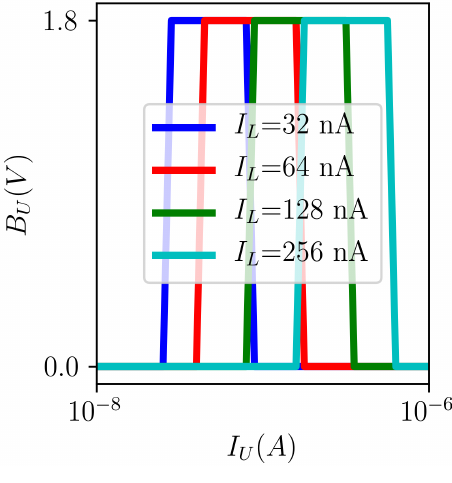}
      \caption{}
      \label{fig:B_IL}
    \end{subfigure}
    \end{center}
    \vspace{-0.2cm}
    \caption{Configurability on the box function. \ref{fig:B_vwell} shows the width of the box function can be varied and \ref{fig:B_vwell} depicts how the box function midpoint can be moved by changing the $I_L$ value in Fig.~\ref{fig:neuralcircuits}.}
  \label{fig:box}
  \vspace{-0.5cm}
\end{figure}
\reffig{fig:box} shows the characteristics of the box function and its configurability using the circuit parameters. In \reffig{fig:B_vwell}, the width of the box is tuned by the well potential shown in \reffig{fig:neuralcircuits} and in \reffig{fig:B_IL}, bias parameter $I_L$ controls the offset of the box function with respect to the normalized sum of the currents from the crossbar array.

\begin{figure*}[t]
  \centering\includegraphics[width=1\textwidth]{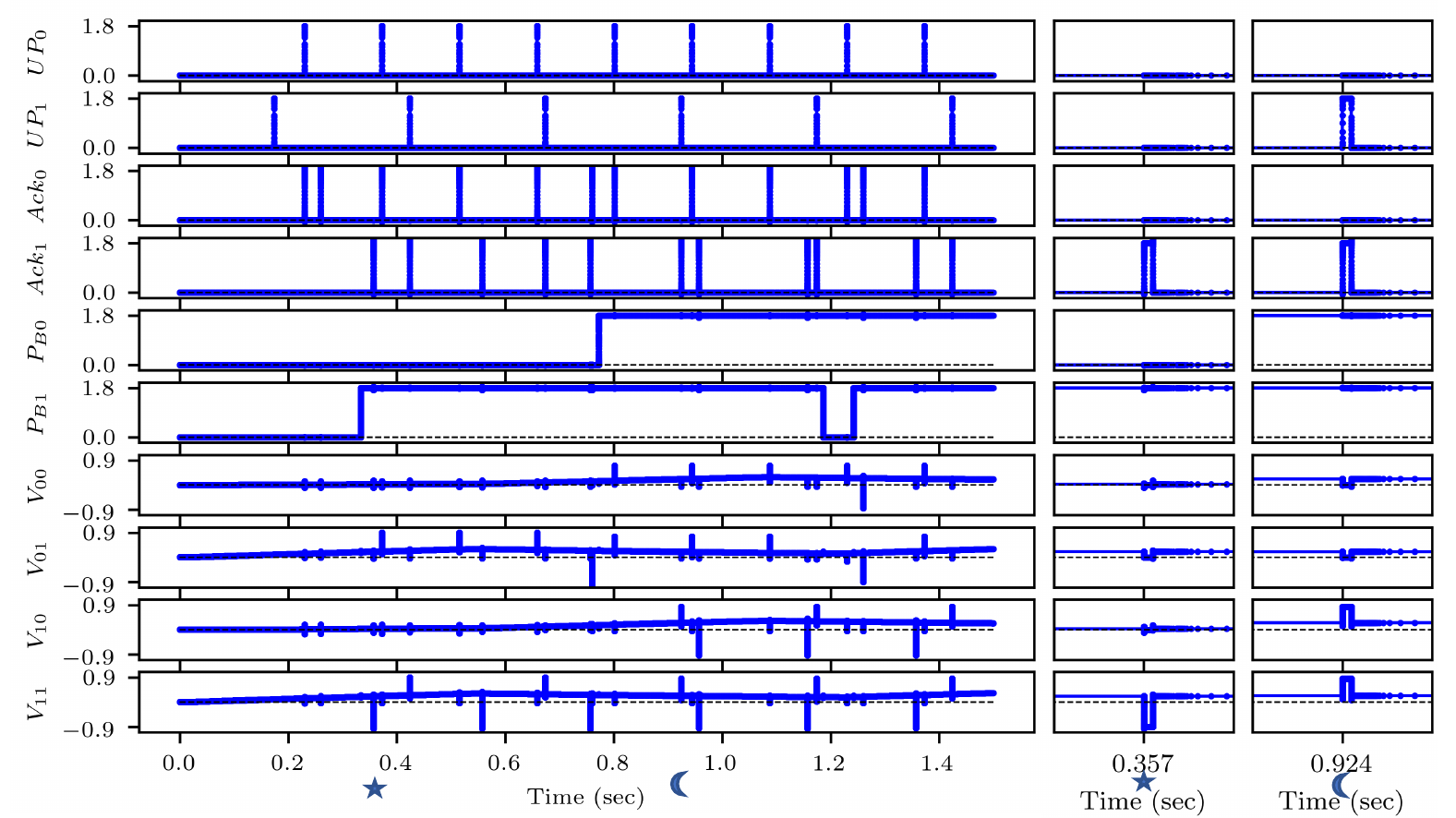}
     \vspace{-0.1in}
  \caption{The learning signals and resulting voltage dropped across the memristive devices. The voltage applied on the device $W_{ji}$ is proportional to $P_i$ in inference and is equal to $V_{set}$ or $V_{rst}$ following the conditions in table~\ref{Update_table}.} 

  \label{fig:circuitresults}
    \vspace{-0.15in}
\end{figure*}

Fig.~\ref{fig:circuitresults} plots learning signals along with the voltages that are dropped across the memristive devices for a $2 \times 2$ array in different scenarios. 
There are two $\delta_s$ signals ($UP_i$ and $DN_i$) with their respective box output ($B_i$) and the output of the learn gate feeding back to the array ($lrn_i)$, and the binary thresholded value of the input signal $P_j$ shown as $\tilde{P}_j$. 
The voltage across the devices match \reftab{Update_table}. 
On the onset of $lrn_i$ signal, if $B_i$ and $P_j$ are non-zero, $V_{set}$ or $V_{rst}$ is applied across the device $W_{ji}$ (in this case 0.9\,V and -0.9\,V respectively), otherwise the voltage across the device is zero. To better illustrate the voltage across the devices, two-time windows are zoomed in and plotted around 0.357\,s and 0.924\,s. 
In the two cases, the $lrn_1$ signal is activated which should only update the devices in the second row. Thus, the voltages $V_{00}$ and $V_{01}$ turn to zero while the $lrn_1$ signal is high. 
In the case of 0.357\,s time window, $DN_1$ is high and $P_0$ and $P_1$ are low and high respectively. Therefore, the voltage across $V_{10}$ is also zero while $V_{11}$ is equal to $V_{rst}$ to decrease the conductance as a result of the { \verb DN } signal. 
In the case of 0.924\,s time window, $UP_1$ is high and $P_0$ and $P_1$ are both high. Therefore, the voltages $V_{10}$ and $V_{11}$ are both equal to $V_{set}$ to increase the conductance as a result of the { \verb UP } signal. 
 
\section{Discussion}
\label{sec:Discussion}

\subsection{Always-on, online learning}
Our architecture supports an always-on learning engine for both inference and learning. 
By default, the \ac{RCA} operates in the inference mode where the devices are read based on the value of $P$ voltages. 
On the arrival of error events, the array briefly enters a learning mode, during which it is blocked for inference. 
During this mode, input events are missed. 
The length of the learning mode depends on the pulse width required for programming the memristive devices, which could be less than $10\,n$s up to $100\,$ns \cite{ielmini2020_review} depending on their type. 
Therefore, based on the frequency of the input events, the maximum size of the array can be calculated. 
The $1T-1R$ memory can be banked with this maximum size. 
The details of this calculation is discussed in section~\ref{sec:hwlim}.

\vspace{-0.2cm}
\subsection{Power and Area Evaluation}
In this subsection, we report the area and power of the learning circuit in our learning architecture. \\
\textbf{Neuronal circuits} The average power and area of the neuronal circuits including the normalizer and box function is estimated to be about 100\,nW and 1000\,$\mu m^2$ respectively. 
\textbf{Spike generator block} The power of this block depends on the time constant of the refractory period which bounds the frequency of the C2F block. If we set the time constant to 10\,ms to limit the frequency to 100\,Hz, the average power consumption of the block is about 10\,uW. The area of the block is about 400\,$\mu m^2$.

\textbf{Filters and \ac{RCA} drivers} The average power and area of these presynaptic circuits including $\tilde{P}$ generation are estimated around $2$\,mW and $3000\,\mu \mathrm{m}^2$, respectively.
We estimated the area and power of the buffer for the case where it can support up to $1$\,mA of current. 
This current is dictated by the size of the array which we discuss in \refsec{sec:dis_scal}.

\vspace{-0.2cm}
\subsection{Scalability}
\label{sec:dis_scal}
\paragraph{Algorithm and architecture}
By proceeding from first principles, namely surrogate gradient descent, this article proposes the design for general-purpose, online \ac{SNN} learning machines. 
The factorization of the learning algorithm as a product of three factors naturally delineates the memory boundaries for distributing the computations. 
In our work, this delineation is realized through \acp{NC} and \acp{PC}.
The separation of the architecture in \acp{NC} and \acp{PC} is consistent with the idea that neural networks are generally stereotypical across tasks, but loss functions are strongly task-dependent. 
The only non-local signal required for learning in an \ac{NC} is the error signal $E$, regardless of which task is learned. 
The ternary nature of the three-factor learning rule and the sparseness afforded by the error-triggering enable frugal communication across the learning data path.

This architecture is not as general as a \ac{GPU}, however, for the following reasons: 1) the \ac{RCA} inherently implements a fully connected network and 2) due to reasons deeply rooted in the spatiotemporal credit assignment problem, loss functions must be defined for each layer, and these functions may not depend on past inputs.  
The first limitation 1) can be overcome by elaborating on the design of the \ac{NC}, for example by mapping convolutional kernels on arrays \cite{Gokmen_etal17_traideep}. There exists no exact and easy solution to the second limitation. However, recent work such as random backpropagation and local learning \cite{N-kland16_direfeed,Mostafa_etal18_deepsupe,Lillicrap_etal16_randsyna, Kaiser_etal20_synaplas,Jaderberg_etal16_deconeur} are likely to provide new strategies to address this limitation in the future.
Finally, although only feedforward weights were trained in our simulations, the approach is fully compatible with recurrent weights as well.

\paragraph{Crossbar array}
Although relatively large arrays of \ac{PCM} and ReRAM devices have been already implemented \cite{sebastian_etal2017_million,azzaz_etal2016_16kb}, nano-scale effects such as IR drop and electromigration may limit the crossbar size. Partitioning the larger layers into smaller memory banks and summing up the output current can be employed as a solution for scaling up \acp{RCA} \cite{shafiee2016isaac, fouda2019mask}.

\paragraph*{Access transistor} 
Since learning is error-triggered, every event can only have one sign and hence for every update, the devices on a row $i$ corresponding to non-zero $\tilde{P}_j$s are updated either to higher or lower conductances together and not both at the same time.
This allows sharing the MUXes at the periphery of the array, making the architecture scalable, since the size of the peripheral circuits grow linearly, while the size of the synapses grows quadratically with the number of neurons. 

\paragraph*{Peripheral circuits}
The size of the $P$ buffer and \ac{TIA} at the end of the row is dependent on the amount of its driving current $I_{drive}$ which is a function of the fan-out $N$. Specifically, in the worst case where all the devices are in their low resistive state, the driving current of the buffer should support:
\[
I_{drive}= N*V_{read}/LRS
\]
where LRS is the low resistive state and $V_{read}$ is the read voltage of the memristive devices. 
Assuming $V_{read}$ of 200\,mV which is a typical value for reading ReRAM and a low resistance of 1\,$k\Omega$ \cite{ielmini2016_review}, in the worst case when all the devices are in their low resistive state, to drive an array with fan-out of 100 neurons, the buffer needs to be able to provide 2\,mA of current. This constraint can be loosened by having a statistic of the weight values in a neural network. For more sparse connectivity this current will drop significantly. 

\vspace{-0.5cm}
\subsection{Error-triggered Learning Impact on Hardware}
\label{sec:energy-eff}
As explained in \refsec{sec:sim_res}, the error-update signals are reduced from 8e6 to 96.7e3 and from 1.3e6 to 14.7e3 for DVSGesture and NMINST, respectively, after applying the error-triggered learning with a small impact of the performance. This reduction is directly reflected on improving the total write energy and lifetime of the memristors with $82.7\times$ and $88.4\times$ for DVSGesture and NMINST, respectively which are considered bottleneck for online learning with memristors \cite{fouda2019spiking}.    
A variant of the error-triggered learning has been demonstrated on the Intel Loihi research chip, which enabled data-efficient learning of new gestures \cite{Stewart_etal20_onlifew-} where learning one new gesture with a DVS camera required only $482 mJ$. Although the Intel Loihi does not employ memristor crossbar arrays, the benefits of error-triggered learning stem from algorithmic properties, and thus extend to the crossbar array.

\section{Conclusion and Future Work}
In this article, we derived a local and ternary error-triggered learning dynamics compatible with crossbar arrays and the temporal dynamics of \acp{SNN}.
The derivation reveals that circuits used for inference and training dynamics can be shared, which simplifies the circuit and suppresses the effects of fabrication mismatch.
By updating weights asynchronously (when errors occur), the number of weight writes can be drastically reduced.
The proposed learning rule has the same computational footprint as error-modulated \ac{STDP} but is functionally different in that there is no acausal part, the updates are triggered on errors if the membrane potential is close to the firing threshold (rather than post-synaptic spike \ac{STDP}). A more detailed comparison of the scaling of this family of learning rules is provided in \cite{Kaiser_etal20_synaplas}.
In addition, the proposed hardware and algorithm can be integrated into spiking sensors such as neuromorphic Dynamic Vision Sensor \cite{delbruck2008frame} to enable energy-efficient computing on the edge thanks to the learning algorithm discussed in \refsec{sec:energy-eff}. 

Despite of the huge benefit of the crossbar array structure, memristor devices suffer from many challenges that might affect the performance unless taken into consideration in the training such as asymmetric non-linearity, precision, and retention \cite{fouda2019spiking}. 
Solutions studied to address these non-idealities, such as training in the loop \cite{Nandakumar_etal20_mixedeep} or adjusting the write pulse properties to compensate them \cite{Fouda_etal19_effeasym}, are compatible with the learning approach proposed in this article. 
Fortunately, on-chip learning helps with other problems such as sneak path (i.e wire resistance), variability, and endurance. 
Combining these solutions and our learning approach will be addressed in future work.   
Interestingly, with error-triggered learning, only selected devices are updated and thus has a direct positive impact on endurance by reducing the number of write events. 
The reduction of write events is directly proportional to the set error rate $\langle | E | \rangle$, and can thus be adjusted based on the device characteristics.
This leads to extending the lifetime of the devices and less write energy consumption.

\section{Acknowledgements}
We acknowledge Giacomo Indiveri for fruitful discussions on the learning circuits.


\begin{IEEEbiography}[{\includegraphics[width=1in,height=1.25in,clip,keepaspectratio]{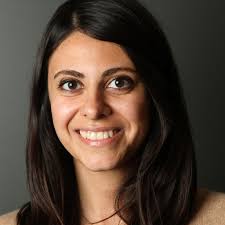}}]{Melika Payvand}
is a research scientist at the Institute of Neuroinformatics, University of Zurich and ETH Zurich. She received her M.S. and Ph.D. degree in electrical and computer engineering from the University of California Santa Barbara in 2012 and 2016 respectively. Her research activities and interest is in exploiting the physics of the computational substrate for online learning and sensory processing.  She is part of the scientific committee of the Capocaccia workshop for neuromorphic intelligence, is serving as a technical member of Neural Systems, Applications and Technologies in Circuits and System society and as a technical program committee for International Symposium on Circuits and Systems (ISCAS). She is a guest editor of Frontiers in Neuroscience and is the winner of the best neuromorph award of the 2019 Telluride neuromorphic workshop.
\end{IEEEbiography}

\begin{IEEEbiography}[{\includegraphics[width=1in,height=1.25in,clip,keepaspectratio]{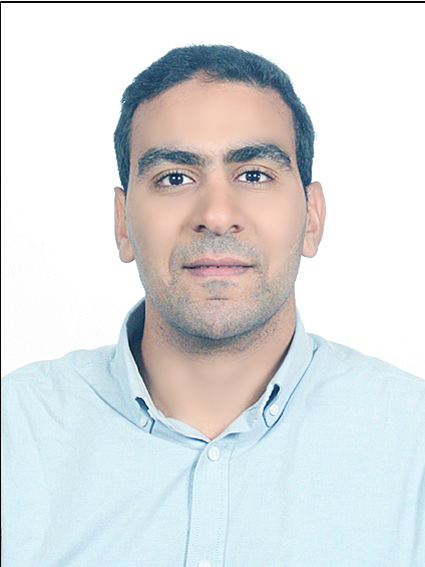}}]{Mohammed E. Fouda}
received the B.Sc. degree (Hons.) in electronics and communications engineering and the M.Sc. degree in engineering mathematics from the Faculty of Engineering, Cairo University, Cairo, Egypt, in 2011 and 2014, respectively. Fouda received his Ph.D. degree from the University of California-Irvine, USA in 2020.  His research interests include brain-inspired computing, neuromorphic circuits and systems, resistive memories, circuit theory, and analog mixed circuits. He serves as peer-reviewer for many prestigious journals and conferences. He also serves as an associate editor in Frontier of Electronics and International Journal of Circuit theory and applications in addition to a technical program committee member in many conferences. He was the recipient of the best paper award in ICM 2013 and the Broadcom foundation fellowship for 2016-2017.
\end{IEEEbiography}
\vspace{-1cm}

\begin{IEEEbiography}[{\includegraphics[width=1in,height=1.25in,clip,keepaspectratio]{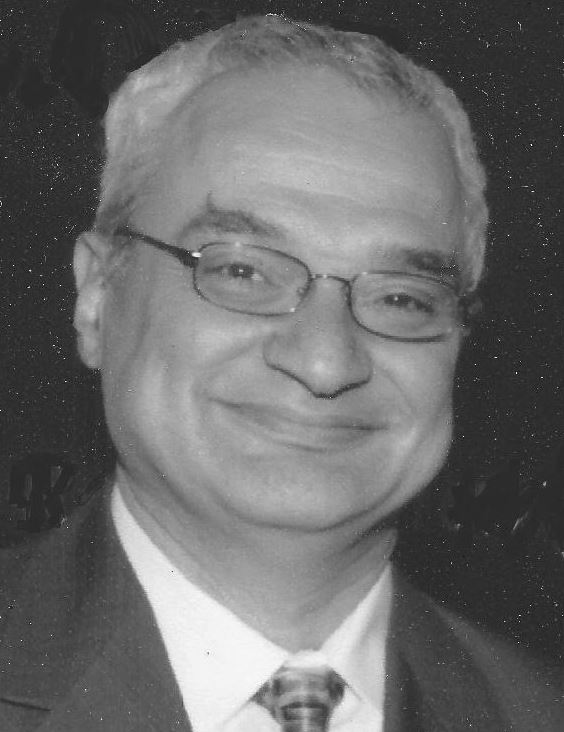}}]{Fadi Kurdahi} (S'85-M'87-SM'03-F'05) ) received the B.E. degree in electrical engineering from the American University of Beirut in 1981 and the Ph.D. degree from the University of Southern California in 1987. Since then, he has been a Faculty with the Department of Electrical Engineering and Computer Science, University of California at Irvine, where he conducts research in the areas of computer-aided design, high-level synthesis, and design methodology of large scale systems, and serves as the Director of the Center for Embedded \& Cyber-physical Systems, comprised of world-class researchers in the general area of Embedded and Cyber-Physical Systems. He is a fellow of the AAAS. He was the Program Chair or the General Chair on program committees of several workshops, symposia, and conferences in the area of CAD, VLSI, and system design. He received the best paper award of the IEEE TRANSACTIONS ON VLSI in 2002, the best paper award in 2006 at ISQED, and four other distinguished paper awards at DAC, EuroDAC, ASP-DAC, and ISQED. He also received the Distinguished Alumnus Award from his Alma Mater, the American University of Beirut, in 2008. He served on numerous editorial boards.
\end{IEEEbiography}
\vspace{-1cm}

\begin{IEEEbiography}[{\includegraphics[width=1in,height=1.25in,clip,keepaspectratio]{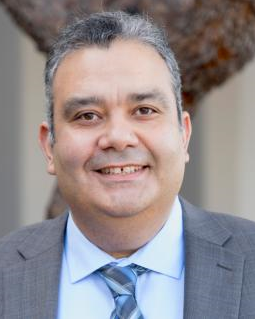}}]{Ahmed E. Eltawil}
(S'97-M'03-SM'14) received the Doctorate degree from the University of California, Los Angeles, in 2003 and the M.Sc. and B.Sc. degrees (with honors) from Cairo University, Giza, Egypt, in 1999 and 1997, respectively. Since 2019 he is a Professor at the Computer, Electrical and Mathematical Science and Engineering Division (CEMSE) at the King Abdullah University of Science and Technology (KAUST), Thuwal, KSA. Since 2005 he was with the Department of Electrical Engineering and Computer Science, at the University of California, Irvine, where he founded the Wireless Systems and Circuits Laboratory. His research interests are in the general area of low power digital circuit and signal processing architectures with an emphasis on mobile systems. Dr. Eltawil has been on the technical program committees and steering committees for numerous workshops, symposia, and conferences in the areas of low power computing and wireless communication system design. He received several awards, as well as distinguished grants, including the NSF CAREER grant supporting his research in low power systems.
\end{IEEEbiography}
\vspace{-1cm}

\begin{IEEEbiography}[{\includegraphics[width=1in,height=1.25in,clip,keepaspectratio]{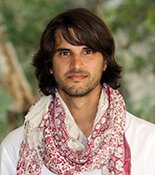}}]{Emre Neftci}
Dr. Emre Neftci received his M.Sc. degree in physics from Ecole Polytechnique Federale de Lausanne, Switzerland, and his Ph.D. in 2010 at the Institute of Neuroinformatics at the University of Zurich and ETH Zurich. Currently, he is an assistant professor in the Department of Cognitive Sciences and Computer Science at the University of California, Irvine. His current research explores the bridges between neuroscience and machine learning, with a focus on the theoretical and computational modeling of learning algorithms that are best suited to neuromorphic hardware and non-von Neumann computing architectures.
\end{IEEEbiography}

\end{document}